\documentclass[pdflatex,sn-mathphys-num]{sn-jnl}

\usepackage{graphicx}%
\usepackage{multirow}%
\usepackage{amsmath,amssymb,amsfonts}%
\usepackage{amsthm}%
\usepackage{mathrsfs}%
\usepackage[title]{appendix}%
\usepackage{xcolor}%
\usepackage{textcomp}%
\usepackage{manyfoot}%
\usepackage{booktabs}%
\usepackage{algorithm}%
\usepackage{algorithmicx}%
\usepackage{algpseudocode}%
\usepackage{listings}%
\usepackage{textcomp}
\usepackage{tikz}
\usepackage{float}
\usepackage[edges]{forest}
\usepackage{tabularray}
\usepackage{rotating}
\usepackage{adjustbox}
\usetikzlibrary{trees}
\usepackage{tablefootnote}
\usepackage{booktabs}
\usepackage{array}
\usepackage{rotating}
\usepackage{adjustbox}
\usepackage{tabularx}
\usepackage{pdflscape}
\usepackage{array}
\usepackage{tikz}
\usepackage{tabularray}
\usetikzlibrary{mindmap}

\theoremstyle{thmstyleone}%
\theoremstyle{thmstyletwo}%

\theoremstyle{thmstylethree}%

\begin{document}

\title[Language Modeling on Tabular Data: A Survey of Foundations, Techniques and Evolution]{Language Modeling on Tabular Data: A Survey of Foundations, Techniques and Evolution}



\author[1,2]{\fnm{Yucheng} \sur{Ruan}}\email{yuchengruan@u.nus.edu}
\equalcont{These authors contributed equally to this work.}
\author[1,2]{\fnm{Xiang} \sur{Lan}}\email{ephlanx@nus.edu.sg}
\equalcont{These authors contributed equally to this work.}

\author[1]{\fnm{Jingying} \sur{Ma}}\email{jingyingma@u.nus.edu}
\author[1]{\fnm{Yizhi} \sur{Dong}}\email{yizhi@nus.edu.sg}
\author[1]{\fnm{Kai} \sur{He}}\email{kai\_he@nus.edu.sg}
\author*[1,2]{\fnm{Mengling} \sur{Feng}}\email{ephfm@nus.edu.sg}

\affil[1]{\orgdiv{Saw Swee Hock School of Public Health}, \orgname{National University of Singapore}, \orgaddress{\country{Singapore}}}

\affil[2]{\orgdiv{Institute of Data Science}, \orgname{National University of Singapore}, \orgaddress{\country{Singapore}}}


\abstract{Tabular data, a prevalent data type across various domains, presents unique challenges due to its heterogeneous nature and complex structural relationships. Achieving high predictive performance and robustness in tabular data analysis holds significant promise for numerous applications. 
Influenced by recent advancements in natural language processing, particularly transformer architectures, new methods for tabular data modeling have emerged. 
Early techniques concentrated on pre-training transformers from scratch, often encountering scalability issues. Subsequently, methods leveraging pre-trained language models like BERT have been developed, which require less data and yield enhanced performance. The recent advent of large language models, such as GPT and LLaMA, has further revolutionized the field, facilitating more advanced and diverse applications with minimal fine-tuning. 
Despite the growing interest, a comprehensive survey of language modeling techniques for tabular data remains absent. This paper fills this gap by providing a systematic review of the development of language modeling for tabular data, encompassing: 
(1) a categorization of different tabular data structures and data types;
(2) a review of key datasets used in model training and tasks used for evaluation;
(3) a summary of modeling techniques including widely-adopted data processing methods, popular architectures, and training objectives;
(4) the evolution from adapting traditional Pre-training/Pre-trained language models to the utilization of large language models;
(5) an identification of persistent challenges and potential future research directions in language modeling for tabular data analysis.
GitHub page associated with this survey is available at:
\url{https://github.com/lanxiang1017/Language-Modeling-on-Tabular-Data-Survey.git}.
}

\keywords{Language modeling, Tabular data, Pre-training language model, Large language model}



\maketitle

\section{Introduction}\label{sec1}

Tabular data, consisting of rows with a consistent set of features, is one of the most prevalent data type in real-world and has been wildly used in different domains \cite{guo2017deepfm, clements2020sequential}. 
In some crucial areas \cite{somani2021deep, bonacin2024exploring, gandhar2024fraud}, a good predictive performance and robustness can provide significant benefits.
However, effectively analyzing tabular data is challenging due to its complex structure. For example, a sample from tabular data could be either a single row of a table (1D tabular data), or a complete table from a set of tables (2D tabular data). Additionally, tabular data typically features a wide range of heterogeneous characteristics \cite{shwartz2022tabular}, such as various data types including numerical, categorical, and textual elements. Furthermore, tables often exhibit complex relationship between both columns and rows.

Over the past few decades, the field of natural language processing (NLP) has witnessed significant evolution in language modeling, particularly with the advent of transformer architecture. 
In the context of tabular modeling, early researches mainly focus on processing tabular data with NLP techniques such as embedding mechanisms, pre-training methods, and architectural modifications. 
These works mainly involve pre-training transformer-based models from scratch specifically for tabular data, which demands a substantial amount of data and can be impractical in certain fields, such as healthcare \cite{gianfrancesco2021narrative, ghebrehiwet2024revolutionizing}. While effective in certain scenarios, these approaches often face challenges in scalability and efficiency. 
Meanwhile, some researchers leverage pre-trained language models (PLMs) (e.g. BERT \cite{devlin2018bert}) to model tabular data. These models, built on top of PLMs, require less training data while delivering superior predictive performance. It shows the effectiveness of adapting and reusing pre-trained LMs on task-specific tabular datasets \cite{howard2018universal}.

More recently, the emergence of large language models (LLMs) has further transformed the landscape. Models such as GPT \cite{brown2020language} and LLaMA \cite{touvron2023llama} have demonstrated remarkable capabilities, achieving state-of-the-art results across a variety of tasks with minimal fine-tuning. These models excel in few-shot and zero-shot learning scenarios, where they can perform complex tasks with little to no additional training data. 
This development has opened new avenues for utilizing LLMs in more advanced and diverse applications for tabular data \cite{radford2019language}.
Strong evidence of this trend is the dramatic increase in the volume of research on tabular modeling with LLMs. This evolution from training models from scratch or using PLMs to adopting LLMs marks a significant paradigm shift in language modeling for tabular data.

Despite significant interest in extracting extensive knowledge from tabular data, there is a notable lack of a comprehensive survey in the research community that clearly sorts out existing language modeling methods on tabular data, outlines technical trends, identifies challenges, and suggests future research directions.
In this work, we bridge this gap by conducting a systematic review of language modeling for tabular data.

\begin{figure}[!h]
    \centering
    \includegraphics[scale=0.4]{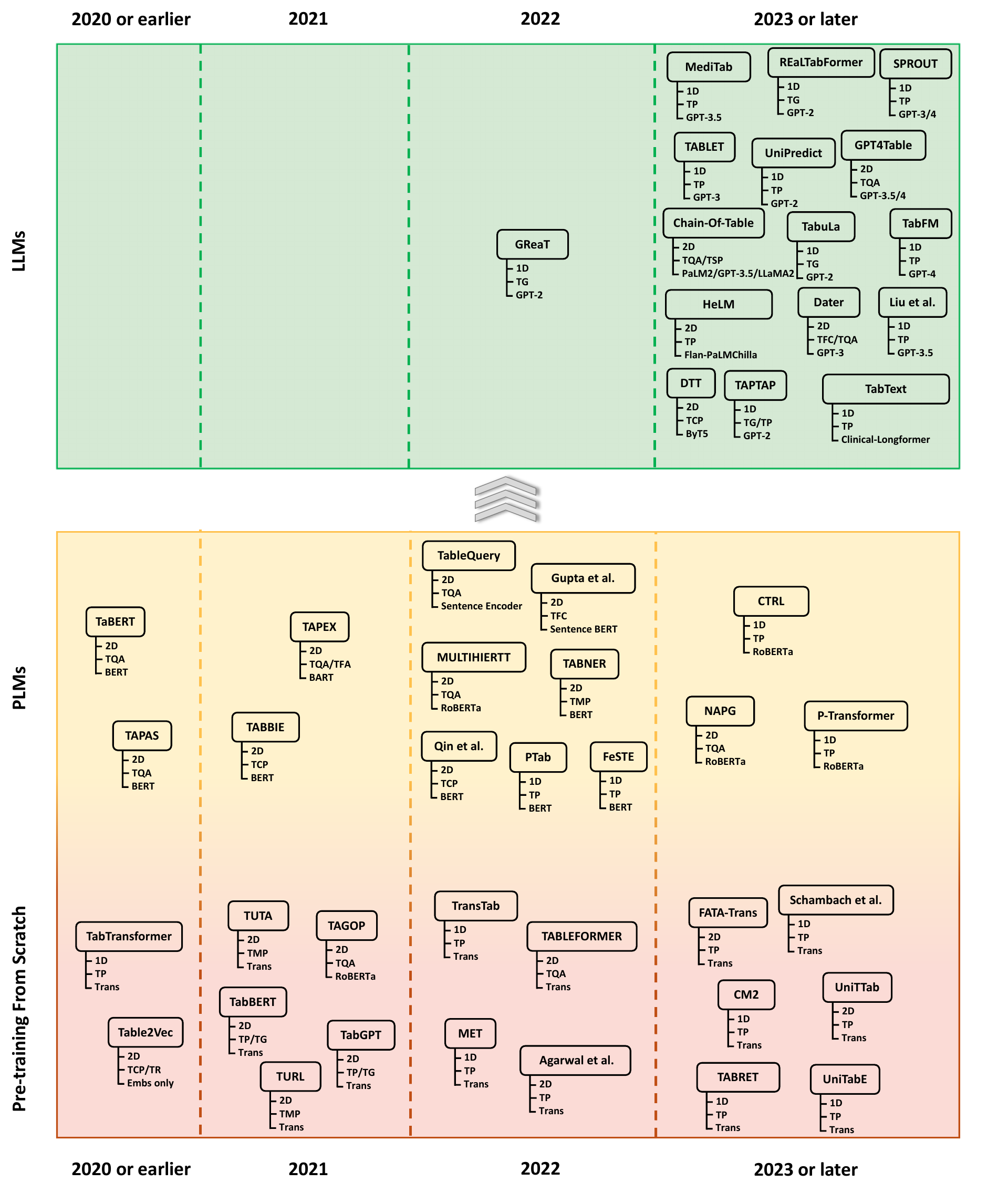}
    \caption{The timeline of the evolution of language modeling on tabular data. Each model includes the following information from top to bottom: type of tabular data (1D or 2D), evaluation tasks, and the backbone model.}
    \label{fig:timeline}
\end{figure}

This survey paper aims to provide a thorough overview of the evolution in language modeling for tabular data (Figure \ref{fig:timeline}) at this pivotal moment of paradigm shift, offering a comprehensive summary and categorization of existing studies to present a clear big picture of this promising research field.

In summary, the main contributions of this survey are threefold.
\textit{First}, it firstly categorizes tabular data into 1D and 2D data formats. Unlike existing surveys either focus on 1D tabular data \cite{borisov2022deep, singh2023embeddings} for traditional tasks like inference and data generation, or concentrate on 2D tabular data \cite{wen2022transformers, badaro2023transformers} for more complex tasks such as information retrieval and table understanding, this work is the first one to provide a systematic review of tasks and datasets for both types of tabular data.
\textit{Second}, it reviews up-to-date progress of language modeling techniques on tabular data and provides an exhaustive taxonomic categorization.
\textit{Third}, it highlights various research challenges and potential avenues for exploration in language modeling for tabular data.

\begin{figure}[!h]
    \centering
    \includegraphics[scale=0.45]{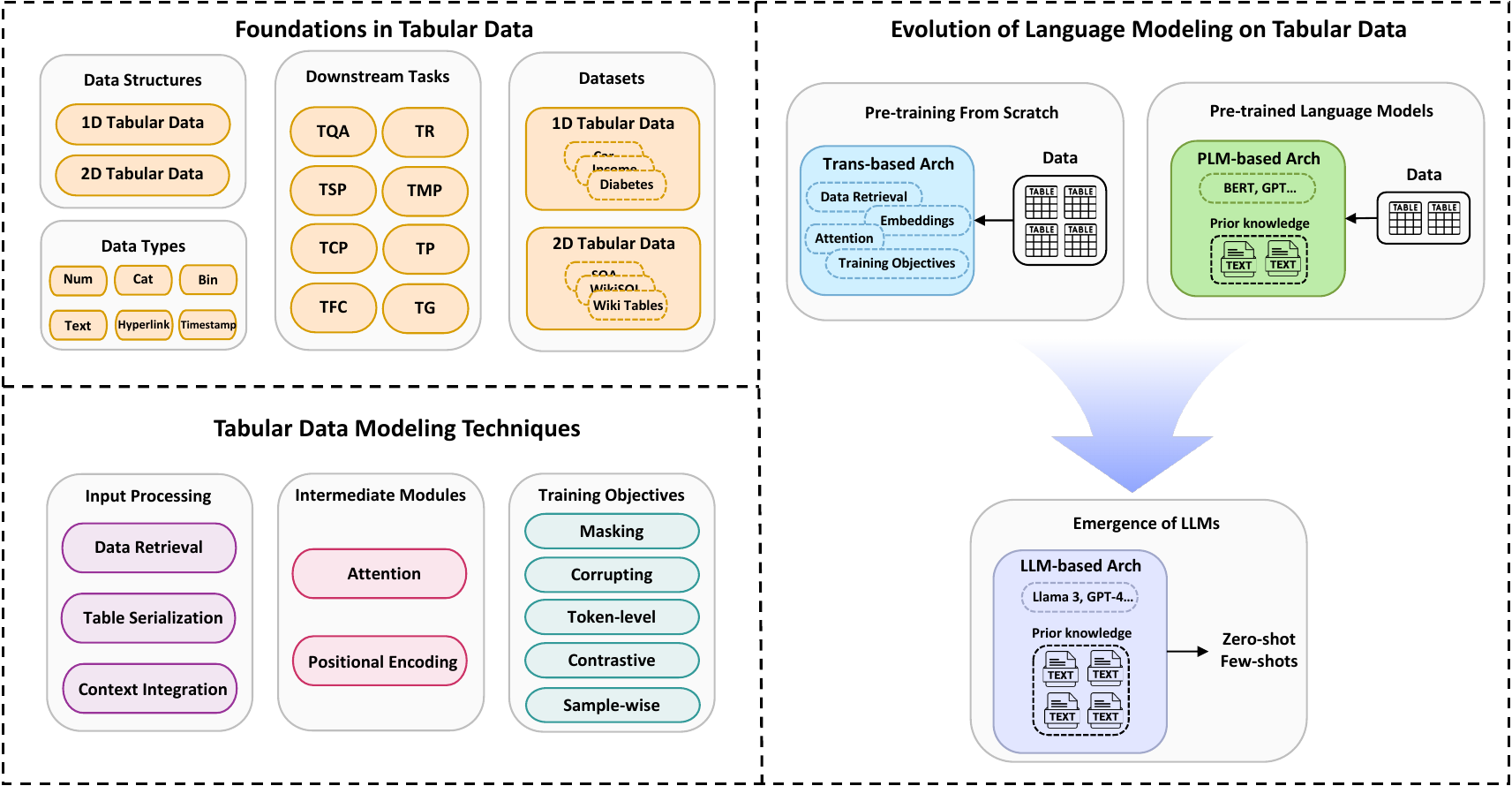}
    \caption{The structure of survey paper. It includes three main parts: foundations in tabular data, tabular data modelling techniques and evolution of language modelling on tabular data.}
    \label{fig:structure}
\end{figure}

In this paper, as illustrated in Figure \ref{fig:structure}, we firstly introduce the foundations of tabular data in Section 2, offering a comprehensive overview of four main parts: data structures (Section 2.1), data types (Section 2.2), downstream tasks (Section 2.3), and datasets (Section 2.4). We explain the two primary tabular data structures that recent research focuses on: 1D and 2D tabular data. We also discuss the different data types found in the tabular domain. Following this, we provide a detailed description of eight major downstream tasks: table question answering (Section 2.3.1), table retrieval (Section 2.3.2), table semantic parsing (Section 2.3.3), table metadata prediction (Section 2.3.4), table content population (Section 2.3.5), table prediction (Section 2.3.6), table fact-checking (Section 2.3.7) and table generation (Section 2.3.8). Subsequently, we outline some commonly used datasets for each data type with key characteristics, which are linked to different downstream tasks.

Next, we present a taxonomy of recent research that maps out the language modeling techniques on tabular data, categorizing it into three key domains: input processing (Section 3.1), intermediate modules (Section 3.2), and training objectives (Section 3.3). Specifically, input processing focuses on transforming the raw tabular data into the format suitable for LMs. We further examine the input processing techniques by breaking them into specific sub-categories: data retrieval (Section 3.1.1), table serialization (Section 3.1.2) and context integration (Section 3.1.3).
In intermediate modules, we discuss two components: positional encoding (3.2.1) and attention mechanisms (3.2.2), which are modified to achieve better predictive performance in tabular domain. Furthermore, we discuss the training objectives, which plays a critical role in helping LMs to learn semantic information.

Following that, we analyze the evolution of how language models are adapted in tabular domain (Section 4). Firstly, we describe the adaptions and benefits of pre-training from scratch and using PLMs early on, particularly with the introduction of transformers (Section 4.1). We then review the recent advance of LLMs in tabular data modeling and highlight how their adaptions differ from previous methods (Section 4.2).

Finally, we identify several challenges and future opportunities in language modeling for tabular data (Section 5), and conclude our paper in Section 6.

\section{Foundations in Tabular Data}
In this section, we discuss the foundations of tabular data, including data structures, related downstream tasks, and common datasets. Specifically, we provide the definition of 1D and 2D tabular data, highlighting the relevant downstream tasks and datasets for each type.

\subsection{Data Structures}
In this survey, we categorize tabular datasets based on their structural characteristics into two main categories. Firstly, 1D tabular datasets usually contain single tables with multiple rows and columns, annotated with task-specific labels. The primary objective of analyzing 1D tabular data is row-level prediction within a single table. In contrast, 2D tabular datasets comprise multiple tables, which may not necessarily be annotated. They are typically leveraged for pre-training or fine-tuning downstream tasks conducted at the table level. 
Subsequently, we present an exploration of various data types for tabular data.

\begin{figure}[!h]
    \centering
    \includegraphics[scale=0.45]{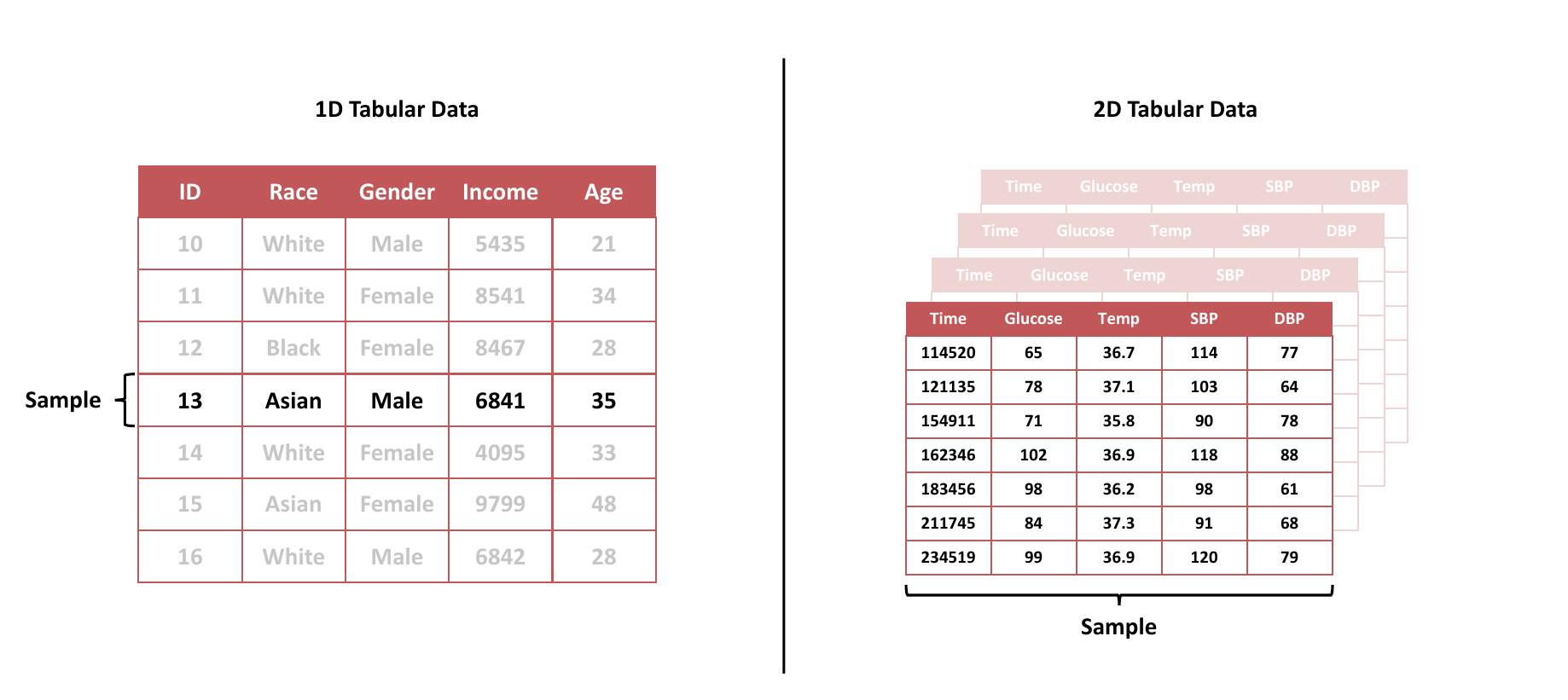}
    \caption{The illustration of 1D tabular data (left) and 2D tabular data (right). One row  represents a sample in 1D tabular data while one tabular table corresponds to a sample in 2D tabular data.}
    \label{fig:tabular_data}
\end{figure}

\subsubsection{1D Tabular Data}
1D tabular data is a common form of structured data. As illustrated in Figure \ref{fig:tabular_data}, it is organized in a specific and well-defined schema, making it easy to query, analyze, and process. It is widely employed for machine learning analysis due to its simplicity in structure. This type of data has a clear and fixed schema, where each data element is organized into rows and columns within a table. A 1D tabular data is defined as $X \in \mathbb{R}^{N \times D} = \left\{x_1, x_2, \ldots, x_D \right\}$, where $D$ is the number of columns and $N$ is the number of rows. Each column $x_d \in \mathbb{R}^{N}$ typically represents a specific attribute or field, and each row $X_i \in \mathbb{R}^{D}$ represents a sample record or entry. Each column adheres to a specific data type constraint, maintaining a straightforward format. Each row within a 1D table incorporates a label according to the specific task, typically a row-level classification or regression based on the sample. Given its simple structure, it is extensively used in classical machine learning models. We define the sample size of 1D tabular data as the number of rows $N$ and feature size as the number of features used as independent variables $D$. 

\subsubsection{2D Tabular Data}
In the modern machine learning era, more complicated tasks have been proposed such as question answering and information retrieval based on table. These tasks involve understanding information from tables and performing table-level analysis. The data used for these tasks are referred to as 2D tabular data. As shown in Figure \ref{fig:tabular_data}, such data comprises a collection of tables. 

Data types in 2D tabular data are more complex (discussed in Section \ref{data_type}). Unlike structured tables where data adheres to column data type constraints, the cells within the 2D tabular data can include diverse types, making the tables to be semi-structured. A 2D tabular data is defined as $X \in \mathbb{R}^{N \times R \times C} = \left\{x_1, x_2, \ldots, x_N \right\}$, where each table $x_i \in \mathbb{R}^{R \times C}$ comprises $R$ rows and $C$ columns,  listing information related to a specific topic. In addition, the dataset often incorporates data from other modalities such as questions, metadata, and associated free-text. The sample size of 2D tabular data is defined as the number of tables $N$.

Furthermore, there is a special type of 2D tabular data, particularly significant for analysis in fields such as healthcare and finance. This data incorporates dynamic temporal information, with timestamps capturing changes over time, thus adding an additional layer of complexity to the dataset's structure. Meantime, information about individuals is recorded across multiple tables, linked by individual identifiers, forming an interconnected 2D data structure.

\subsection{Data Types} \label{data_type}
The values of individual cells within tabular data are governed by the constraints of column data types, ensuring regularity and consistency. Commonly, 1D datasets include simple data types such as numerical, categorical, and binary data. In contrast, the data types in 2D tabular data are characterized by greater flexibility and complexity. In the following section, we elaborate on the diverse data types that emerge in tabular data.
\begin{itemize}
    \item \textit{Numerical data.} Numerical data consists of values that can be expressed as numbers. The values can either be discrete, such as integers denoting whole numbers, or continuous, such as floating-point numbers with decimal precision. Numerical data type is a fundamental data type, allowing direct mathematical calculation.
    \item \textit{Categorical data.} Categorical data, also known as qualitative data, represents data that can be grouped into a finite set of distinct categories. These categories can be nominal without any order information, or ordinal, exhibiting sequential rankings, albeit inter-category distances are unknown.
    \item \textit{Binary data.} Binary data is a special case of categorical data where each observation falls into one of two distinct categories. These categories are often labeled as “Yes” and “No”, represented numerically as 1 and 0, or expressed as Boolean values, representing logical outcomes denoted by “True” and “False”.
    \item \textit{Text data.} Text data includes both structured texts and free-texts. Structured texts follow a predefined format and a structured schema, such as name and address. On the other hand, free-texts refer to unstructured textual data that lacks a standardized structure. They comprise collections of text without format limitations, such as comments and descriptions.
    \item \textit{Hyperlinks.} Hyperlinks contained in tabular data serve as connections to external sources, such as websites or documents, providing the dataset with supplementary context. The presence of hyperlinks in tabular data necessitates multi-hop reasoning over both the tables and the hyperlinked contents.
    \item \textit{Timestamps.} Timestamp data denote the date and time information when an event occurs. By providing information on the sequence and intervals of events, it plays a pivotal role in tracking events over time. Capturing temporal information enables the analysis of time-series data, offering valuable insights into trends and patterns.
\end{itemize}

\subsection{Downstream Tasks}
With the advent and widespread deployment of language models, particularly LLMs, there has been a expansion in the spectrum of downstream tasks related to tabular data modeling using language models. 
These tasks are also executed with higher accuracy using LLMs compared to previous models. 
In this section, we'll elaborate 8 distinct categories of downstream tasks for which language models are employed. A summary of downstream tasks are provided in Table \ref{tab:downstream_task}. We provide a comprehensive overview of these tasks, detailing their nature, the inputs they require, the outputs they generate and specific cases. Also, we show the evaluation metric for each task in Table \ref{tab:metric}.

\subsubsection{Table Question Answering (TQA)}
In the Table Question Answering (TQA) task, the system answers questions based on information contained in structured tables. Since TQA task requires understanding both the question (natural language) and the table's schema to provide accurate answers, language models could help. TQA is divided into simple TQA and complex TQA \cite{badaro2023transformers}. Simple TQA involves looking for answers within the table's cells, while complex TQA involves aggregation or reasoning based on the table's information to get the answer. The input of a TQA task is a question and a table, while the output is the answer to the question. 

The evaluation metrics for the TQA task mainly include Execution Accuracy, Exact Match (EM), and F1. Execution Accuracy measures the model's ability to generate correct answers given a table and a question. It calculates accuracy by executing the query generated by the model and comparing the results with the expected query results. This metric was used by \cite{yin-etal-2020-tabert} \cite{zhang2023napg} for evaluation in their paper. EM evaluates the extent to which the model's generated answer matches the standard answer exactly. Specifically, Exact Match requires that the content and format of the model's generated answer be identical to the standard answer. This metric was used by \cite{zhang2023napg} \cite{zhao2022multihiertt} in their paper.

\subsubsection{Table Retrieval (TR)}
Table Retrieval (TR) involves extracting relevant information from multiple tables based on a given query. The input is a question, and the output is the most relevant cells or tables. Language models can significantly aid this task by understanding the context of the query and the semantics of the data, enabling more accurate and efficient retrieval.

The evaluation metrics for the Table Retrieval task mainly include Normalized Discounted Cumulative Gain (NDCG). NDCG is a commonly used metric to measure the ranking effectiveness of information retrieval systems. It calculates the cumulative gain of the retrieval results, applies a discount to the gain, and normalizes it to evaluate the system's ability to return relevant results at various positions. NDCG was first introduced by \cite{Webber2002} and has been widely used. For example, Table2vec \cite{zhang2019table2vec}, RIM \cite{qin2021retrieval}, PET \cite{du2022learning}, Table2Graph \cite{zhou2022table2graph} used NDCG for evaluation in their proposed methods.

\subsubsection{Table Semantic Parsing (TSP)}
Table Semantic Parsing (TSP), mainly includes Text to SQL, which translates natural language queries into SQL commands. The input is a user's textual query, and the output is an executable SQL query. Language models could assist by understanding the textual query first, and then mapping it accurately to SQL syntax. 

The evaluation metrics for the Table Semantic Parsing task mainly include Accuracy and Logical Form Accuracy (AccLF). Accuracy measures whether the answers generated by the model match the standard answers. For example, TAPEX \cite{liu2021tapex} and StructGPT \cite{jiang2023structgpt} used Accuracy for evaluation in their paper. AccLF evaluates whether the logical forms generated by the model match the standard logical forms. For example, Seq2SQL \cite{Zhong2017} detailed discussed the application of AccLF in SQL query generation.

\subsubsection{Table Metadata Prediction (TMP)}
Table Metadata Prediction (TMP) aims at inferring various types of metadata about tables, which includes subtasks such as Column Type Annotation. The input to this task is typically a row or a semi-structured table, and the output is enriched metadata information: Column Type Prediction categorizes each column by their data types ($e.g.$, integer, text, date); Table Type Classification determines the inherent property of the table ($e.g.$, transactional, relational, or summary tables). TMP could also involves identifying and understanding the relationships within and between tables and their entities (typically refers to the cell value and elements within a table, which could be specific data in a cell value such as names, places, dates, numbers, etc., or could also refer to elements of the table such as rows or columns), encompassing subtasks such as Table Relation Prediction, Table Relation Extraction, Table Entity Linking, Entity-based Relation Retrieval, and Cell Entity Recognition. The input for this task includes one or more tables along with potential external knowledge sources such as knowledge graph. The output ranges from identified relationships between tables to linked entities within cells to external knowledge sources, and evaluations of how entities and their relationships are represented within a table. Language models could significantly aid in these tasks by leveraging their ability to to process and interpret natural language and to understand context and semantics from textual data. For example, in column type annotation task, language models understand linguistic nuances and naming conventions in table headers. For instance, a language model can discern that a header containing "Date Of Birth" likely refers to a date type, whereas "Total Amount" or "Quantity" suggests numeric data. This nuanced understanding stems from the model's training on a vast corpus of text, enabling it to capture a broader range of expressions and context than typical data mining approaches. 

The common evaluation metrics for the sub-task Column Type Annotation are Precision, Recall, F1, and Macro-F1. Additionally, the common evaluation metric for the sub-tasks Table Type Classification and Cell Entity Recognition is F1. Furthermore, the common evaluation metrics for the tasks Table Relation Extraction and Table Entity Linking are F1, Precision, and Recall. Finally, the common evaluation metrics for the sub-tasks Table Relation Prediction and Entity-based Relation Retrieval are Accuracy at 1, 3, and 5 (Precision at 1, 3, 5, and 10). Accuracy/Precision at $n$ is a metric used to evaluate the performance of a model in recommendation or retrieval tasks. Accuracy at $n$ indicates whether the model includes the target answer within the top $n$ recommended positions. It measures the accuracy of the target answer being within the top 1 to $n$ positions of the recommendation list. Precision at $n$ refers to the frequency with which the model accurately recommends the target answer within the top $n$ positions of the recommendation list. For example, Qin el al \cite{qin2022external} used Accuracy at $n$ for evaluation.

\subsubsection{Table Content Population (TCP)}
The aim of Table Content Population (TCP) is enriching and completing tables by addressing subtasks such as Column Population, Row Population, and Cell Filling. The input for this task is a table that may have missing or incomplete information in its columns, rows, or cells. The output is a more complete table, with all missing elements accurately filled in. Language models can assist with these tasks by leveraging their knowledge base and understanding of context to infer missing information accurately.

The evaluation metrics for the sub-tasks Column Population and Row Population are Mean Average Precision (MAP), Mean Reciprocal Rank (MRR), Re-call, Normalized Discounted Cumulative Gain at 10 (NDCG-10), Normalized Discounted Cumulative Gain at 20 (NDCG-20). MAP measures the average precision of a model across multiple queries. It calculates the average precision (AP) for each query and then takes the mean of these APs. MAP effectively evaluates the overall performance of a model in handling multiple queries. For example, Table2Vec \cite{zhang2019table2vec} used MAP for evaluation. MRR assesses the rank of the first correct answer returned by the model. It calculates the reciprocal rank for each query and then averages these values. A higher MRR indicates that the model returns the correct answer earlier. For exapmle, RIM \cite{qin2021retrieval} used MRR for evaluation. Recall measures the proportion of all relevant answers that the model can identify. NDCG-n evaluates the ranking effectiveness of a model in the top n recommended positions. By applying a discount to the cumulative gain at the top n positions and normalizing it, NDCG-n assesses the model's ability to return relevant results at higher ranks. For example, PET \cite{du2022learning} used NDCG-5 and NDCG-10 for evaluation. The common evaluation metrics for the sub-task Cell Filling is Precision at $n$.

\subsubsection{Table Prediction (TP)}
Table Prediction (TP) is the most common task in the field of tabular modeling, incorporating subtasks like regression and classification. The input is a dataset organized in table format, featuring rows of instances and columns representing features or variables. For regression, the output is a continuous value, whereas for classification, the output categorizes each instance into discrete classes or labels. Since tables can contain cells with free-text, language models could significantly aid in these tasks by leveraging their vast pre-trained knowledge and contextual understanding.

The evaluation metrics for classification include F1 score, AUROC, and Accuracy, while for prediction tasks, they encompass Root Mean Square Error (RMSE), Mean Absolute Error (MAE), and $R^2$. RMSE quantifies the average magnitude of errors between predicted and actual values. MAE similarly measures the average of absolute errors in predictions. The $R^2$ statistic, also known as the coefficient of determination, tells us what percentage of the variation in the outcome can be explained by the model.

\subsubsection{Table Fact-checking (TFC)}
Table Fact-checking (TFC) is to verify the accuracy and truthfulness of a statement based on knowledge presented in tables. The input is a table, alongside claims or statements. The output is a quantification of the veracity of these claims, categorizing them as true, false, or partially true/false based on the evidence within the table. Language models can be helpful in this task due to their ability to process natural language and their knowledge in pre-trained models. The common evaluation metric for TFC is Accuracy.

\subsubsection{Table Generation (TG)}
Table Generation is the process of creating structured tables from given inputs, which can range from specific data points to natural language descriptions or queries. The output is a well-structured table that organizes the input data into columns and rows, effectively presenting the information in a clear, concise, and accessible format. This task was born with the emergence of large language models.

The evaluation metrics for the Table Generation task mainly include Bilingual Evaluation Understudy (BLEU) and Average Normalized Edit Distance (ANED). BLEU is a metric used to assess the quality of generated text, initially designed for machine translation tasks. It calculates scores by comparing the n-gram matches between the generated text and multiple reference texts. The score ranges from 0 to 1, with a score closer to 1 indicating a higher similarity between the generated text and the reference texts. This metric was first proposed by Papineni et al. \cite{Papineni2002} in their paper. UniTabE \cite{yang2023unitabe}, GPT4Table \cite{sui2023gpt4table} and DATER \cite{ye2023large} used BLEU for evaluation in their paper. ANED is a metric used to evaluate the edit distance between the generated text and reference texts. It calculates the number of edit operations (such as insertions, deletions, and substitutions) needed to transform the generated text into the reference text, then normalizes the score to a range from 0 to 1. A lower score indicates a higher similarity between the generated text and the reference text. This metric was first proposed by Marzal et al. \cite{Marzal1993} in their paper. DTT \cite{nobari2023dtt} used ANED for evaluation in their paper.

\begin{table}
\centering
\begin{talltblr}[
  caption = {Summary of downstream tasks related to tabular data.},
  label = {tab:downstream_task},
]{
  width = \linewidth,
  colspec = {Q[50]Q[200]Q[150]Q[150]Q[150]},
  row{1} = {c},
  cell{2}{1} = {r = 2}{},
  cell{2}{3} = {r = 2}{},
  cell{6}{1} = {r = 7}{},
  cell{13}{1} = {r = 3}{},
  cell{16}{1} = {r = 2}{},
  cell{20}{1} = {c = 6}{0.934\linewidth},
  hline{1-2,4-6,13,16,18-20} = {-}{},
}
\textbf{Task} & \textbf{Sub-tasks} & \textbf{Input} & \textbf{Output} & \textbf{Exapmle} \\

TQA & Simple QA & Table and context  & Cell value & TaBERT \cite{yin-etal-2020-tabert} \\ 

& Complex QA & & Aggregation Prediction & TableQAKit \cite{lei2023tableqakit} \\

TR & - & Table and query & Tables or cells with ranking scores & Table2vec \cite{zhang2019table2vec} \\

TSP & Text to SQL & Table and query & SQL sentence & TAPEX \cite{liu2021tapex} \\

TMP & Column Type Annotation & Table and a set of semantic types & Table type & TABBIE \cite{iida2021tabbie} \\

& Table Type Classification & Table & Table type & TUTA \cite{iida2021tabbie} \\

& Table Relation Prediction & 2 colum without header & Relations & Qin el al \cite{qin2022external} \\

& Table Relation Extraction & Table and a set of relations R in knowledge base & Relations & TURL \cite{deng2022turl} \\

& Table Entity Linking & Table and Knowledge base & Entities Candidates & TURL \cite{deng2022turl} \\

& Entity-based Relation Retrieval & 2 cells and Knowledge Graph & Relations & Qin el al \cite{qin2022external} \\

& Cell Entity Recogonization & 2 cells & Relations & TABNER \cite{koleva2022named} \\

TCP & Column Population & The first N columns of a table & Cells value & Table2vec \cite{zhang2019table2vec}, TABBIE \cite{iida2021tabbie} \\

& Row Population & Table and set of entities & Entities & Table2vec \cite{zhang2019table2vec}, TURL \cite{deng2022turl} \\

& Cell Filling & Partial Tables & Object entity & TURL \cite{deng2022turl} \\

TP & Classification & Table & Category & CTRL \cite{li2023ctrl}, CT-BERT \cite{ye2023ct}, ~\textbf {MediTab \cite{wang2023meditab}}, \textbf {UniPredict \cite{wang2023unipredict}} \\

& Regression & Table & Probability & DANets \cite{chen2022danets}, \textbf{TAPTAP} \cite{zhang2023generative} \\

TFC & - & Table and statement & Binary decision & TAPEX \cite{liu2021tapex}, \textbf{gpt4table} \cite{nam2023semi} \\

TG & - & Text description & Table & \textbf{GReaT} \cite{nam2023semi}, \textbf{TabuLa} \cite{zhao2023tabula}, \textbf{REaLTabFormer} \cite{solatorio2023realtabformer}\\

Note: In the \textbf{Example} column, the bolded entries are Large Language Models, and the non-bolded entries are Language Models. &&&&&

\end{talltblr}
\end{table}

\begin{table}
\centering
\begin{talltblr}[
  caption = {Summary of metrics for tabular downstream tasks.},
  label = {tab:metric},
]{
  width = \linewidth,
  colspec = {Q[50]Q[200]Q[400]},
  row{1} = {c},
  cell{2}{1} = {r = 2}{},
  cell{2}{3} = {r = 2}{},
  cell{6}{1} = {r = 7}{},
  cell{7}{3} = {r = 2}{},
  cell{9}{3} = {r = 2}{},
  cell{11}{3} = {r = 2}{},
  cell{13}{1} = {r = 3}{},
  cell{13}{3} = {r = 2}{},
  cell{16}{1} = {r = 2}{},
  hline{1-2,4-6,13,16,18-20} = {-}{},
  hline{7, 9, 11, 15} = {2-3}{solid},
}
\textbf{Task} & \textbf{Sub-tasks} & \textbf{Metrics} \\

TQA & Simple QA & Execution accuracy, Exact Match (EM), F1 \\ 

& Complex QA &  \\

TR & - & Normalized Discounted Cumulative Gain (NDCG) \\

TSP & Text to SQL & Accuracy, Logical Form Accuracy (AccLF) \\

TMP & Column Type Annotation & Precision, Recall, F1, Macro-F1 \\

& Table Type Classification & F1 \\

& Cell Entity Recogonization & \\

& Table Relation Extraction & F1, Precision, Recall \\

& Table Entity Linking & \\

& Table Relation Prediction & Accuracy at $n$ (Precision at $n$) \\

& Entity-based Relation Retrieval &  \\

TCP & Column Population & Mean Average Precision (MAP), Mean Reciprocal Rank (MRR), Recall, Normalized Discounted Cumulative Gain at $n$ (NDCG-$n$)\\

& Row Population & \\

& Cell Filling & Precision at $n$ \\

TP & Classification & F1, AUROC, Accuracy \\

& Regression & Root Mean Square Error (RMSE), Mean Absolute Error (MAE), $R^2$ \\

TFC & - & Accuracy \\

TG & - & Bilingual Evaluation Understudy (BLEU), Average Normalized Edit Distance (ANED) \\

\end{talltblr}
\end{table}

\subsection{Datasets}
In the scope of this survey, 1D tabular data were extensively employed during the fine-tuning stage of language models with supervised TP downstream tasks, which enables the adjustment of the learned representation according to the specific task. They can also be used to evaluate the performance of tabular data mining models. Table \ref{tab:dataset_1d} presents an overview of the most prevalent 1D datasets identified in the surveyed literature. The 1D dataset spans various domains including census, financial, and healthcare sectors, including regression, classification, and top-n ranking sub-tasks. The sample size varies from a few hundred to 11 million, with feature sizes typically remaining under 50, although notable exceptions such as MNIST  \cite{deng2012mnist} (784 features) and BlogFeedback \cite{misc_blogfeedback_304} (280 features). One popular example is the UCI adult income (Income) dataset \cite{misc_adult_2}, it is a well-known dataset that contains census variables such as age, work class, and education level. This dataset constitutes a binary classification dataset, with labels denoting whether income exceeds \$50K/yr. Additionally, the diabetes dataset \cite{smith1988using} from OpenML presents another binary classification challenge, aimed at predicting the onset of diabetes mellitus using demographic and clinical attributes.

\begin{table}
\small
\centering
\begin{talltblr}[
  caption = {Summary of 1D tabular datasets including prediction sub-tasks, sample size, feature size and models using the datasets.},
  label = {tab:dataset_1d},
  note{1} = {https://www.kaggle.com/datasets/blastchar/telco-customer-churn}
]{
  width = \linewidth,
  colspec = {Q[160]Q[120]Q[70]Q[70]Q[350]},
  row{1} = {c, m},
  hline{1-2,15} = {-}{},
}
\textbf{Dataset}                                      & \textbf{Prediction Sub-Task} & \textbf{Sample Size} & \textbf{Feature Size} & \textbf{Example}                                                                                                                   \\
Income \cite{misc_adult_2}                            & Classification    & 48,842               & 14                    & STab \cite{hajiramezanali2022stab}, TABLET \cite{slack2023tablet}, TabuLa \cite{zhao2023tabula}                                    \\
Diabetes \cite{smith1988using}                        & Classification    & 768                  & 8                     & STUNT \cite{nam2023stunt}, TABRET \cite{onishi2023tabret}, TAPTAP \cite{zhang2023generative}                                       \\
Forest Cover Type \cite{misc_covertype_31}            & Classification    & 581,012              & 52                    & DANets \cite{chen2022danets}, TabNet \cite{arik2021tabnet}, DANets, MET \cite{majmundar2022met}                                    \\
California Housing \cite{pace1997sparse}              & Regression        & 20,640               & 8                     & Gorishniy et al. \cite{gorishniy2022embeddings}, GReaT \cite{borisov2022language}, REaLTabFormer \cite{solatorio2023realtabformer} \\
MNIST (tabular) \cite{deng2012mnist}                  & Classification    & 7,000                & 784                   & SDAT \cite{fang2022semi}, MET \cite{majmundar2022met}, Contrastive Mixup \cite{darabi2021contrastive}                              \\
Qsar Bio \cite{misc_qsar_biodegradation_254}          & Classification    & 1,055                & 41                    & PTab \cite{liu2022ptab}, TabPFN \cite{hollmann2022tabpfn}, Schambach et al. \cite{schambach2023scaling}                            \\
Credit-g \cite{misc_statlog_(german_credit_data)_144} & Classification    & 1,000                & 20                    & UniTabE \cite{yang2023unitabe}, TabPFN \cite{hollmann2022tabpfn}, Liu et al. \cite{liu2023investigating}                           \\
Higgs Boson \cite{misc_higgs_280}                     & Classification    & 11M                  & 28                    & TabNet \cite{arik2021tabnet}, CT-BERT \cite{ye2023ct}, Schambach et al. \cite{schambach2023scaling}                                \\
BlogFeedback \cite{misc_blogfeedback_304}             & Regression        & 60,021               & 280                   & VIME \cite{yoon2020vime}, SubTab \cite{ucar2021subtab}, SDAT \cite{fang2022semi}                                                  \\
Bank Marketing \cite{misc_bank_marketing_222}         & Classification    & 45,211               & 16                    & TabTransformer \cite{huang2020tabtransformer}, TabLLM \cite{hegselmann2023tabllm}, PTab \cite{liu2022ptab}                         \\
BlastChar\TblrNote{1}                                             & Classification    & 7,043                & 20                    & TabTransformer \cite{huang2020tabtransformer}, SAINT \cite{somepalli2021saint}, PTab \cite{liu2022ptab}                            \\
MovieLens-1M \cite{harper2015movielens}               & Top-n ranking     & 1M                   & 7                     & RIM \cite{qin2021retrieval}, PET \cite{du2022learning}, Table2Graph \cite{zhou2022table2graph}                                     \\
Car \cite{misc_car_evaluation_19}                     & Classification    & 1,728                & 6                     & TabLLM \cite{hegselmann2023tabllm}, SPROUT \cite{nam2023semi}, CT-BERT \cite{ye2023ct}                                     
\end{talltblr}
\end{table}

Table \ref{tab:dataset_2d_task} summarizes the downstream tasks and sample sizes of the most prevalent 2D tabular datasets identified, where 'x' indicates the tasks applicable to each dataset. The majority of 2D datasets are applicable or designed for TQA. Table \ref{tab:dataset_2d_info} demonstrates the table metadata, context information, and task related context associated with each dataset. 2D tabular datasets can be either unlabeled or labeled. Unlabeled datasets such as Wikipedia Tables and WDC Web Table Corpus \cite{lehmberg2016large} are vast collections of tables, each containing information on a specific topic. They are often used for pre-training to facilitate the language model’s understanding of semantics and knowledge of tabular data. In contrast, labeled datasets with task-specific annotations are used during the fine-tuning stage. For instance, the WikiTableQuestion dataset \cite{pasupat2015compositional} comprises table-associated questions and corresponding answers. Data types within 2D tabular data can be more complicated, which requires more nuanced reasoning. For example, in MULTIHIERTT \cite{zhao2022multihiertt}, each document contains multiple tables, which often exhibit hierarchical structures with multi-level headers. Tables in HybridQA \cite{chen2020hybridqa} incorporate hyperlinks leading to Wikipedia passages, requiring multi-hop reasoning. Furthermore, tables within the dataset can be inter-related. In Spider \cite{yu2018spider}, the task involves joining multiple tables within a database by keys. 

\begin{table}
\small
\centering
\begin{talltblr}[
  caption = {Properties of 2D tabular datasets including downstream tasks and sample size.},
  label = {tab:dataset_2d_task},
  note{1} = {WikiTable: \href{https://github.com/bfetahu/wiki_tables}{https://github.com/bfetahu/wiki\_tables}}
]{
  width = \linewidth,
  colspec = {Q[120]Q[32]Q[32]Q[32]Q[32]Q[32]Q[32]Q[55]},
  row{1} = {c},
  row{2} = {c},
  cell{1}{1} = {r=2}{},
  cell{1}{2} = {c=6}{},
  cell{1}{8} = {r=2}{},
  hline{1,3,15} = {-}{},
  hline{2} = {2-7}{},
}
\textbf{Dataset}                 & \textbf{Downstream Task} &              &              &             &              &              & \textbf{Sample Size}  \\
                                                  & \textbf{TFC}   & \textbf{TQA} & \textbf{TSP} & \textbf{TR} & \textbf{TMP} & \textbf{TCP} &                       \\
Wikipedia Tables\TblrNote{1}                                  &                & x            & x            & x           & x            & x            & \textasciitilde{}2.62M       \\
WDC Web Table Corpus \cite{lehmberg2016large}     &                & x            & x            &             & x            &              & \textasciitilde{}233M \\
WikiTableQuestion \cite{pasupat2015compositional} &                & x            & x            &             &              &              & 2,108                 \\
WikiSQL \cite{zhong2017seq2sql}                   &                & x            & x            & x           &              &              & 24,241                \\
Spider \cite{yu2018spider}                        &                &              & x            & x           &              &              & 1,020                 \\
SQA \cite{iyyer2017search}                        &                & x            & x            &             &              &              & 982                   \\
MULTIHIERTT \cite{zhao2022multihiertt}            &                & x            &              &             &              &              & \textasciitilde{}9.8K \\
FeTaQA \cite{nan2022fetaqa}                       &                & x            & x            &             &              &              & 10,330                \\
TAT-QA \cite{zhu2021tat}                          &                & x            &              &             &              &              & 2,757                 \\
HybridQA \cite{chen2020hybridqa}                  &                & x            &              &             &              &              & 13,000                \\
ToTTo \cite{parikh2020totto}                      &                & x            &              &             &              &              & 83,141                \\
TabFact \cite{chen2019tabfact}                    & x              &              &              &             &              &              & 16,573                
\end{talltblr}
\end{table}

\begin{sidewaystable}
\small
\centering
\begin{talltblr}[
  caption = {Summary of context of 2D tabular datasets with models using the datasets.},
  label = {tab:dataset_2d_info},
  note{1} = {WikiTable: \href{https://github.com/bfetahu/wiki_tables}{https://github.com/bfetahu/wiki\_tables}}
]{
  width = \linewidth,
  colspec = {Q[130]Q[350]Q[180]},
  row{1} = {c},
  hline{1-2,14} = {-}{},
}
\textbf{Dataset}                                  & \textbf{Other Modalidities}                                                                                                                                                                                                                   & \textbf{Example}                                                                                                 \\
Wikipedia Tables\TblrNote{1}                                  & \textbf{Table Metadata:} titles, captions, and NL contexts                                                                                                                                                                                                                & Table2Vec \cite{zhang2019table2vec}, TURL \cite{deng2022turl}, TABBIE \cite{iida2021tabbie}                      \\
WDC Web Table Corpus \cite{lehmberg2016large}     & {\textbf{Table Metadata:} table orientation, header rows, key columns\\\textbf{Context Information:} the title of the HTML page, the caption of the table, the text before and after the table, and timestamps from the page}                 & TaBERT \cite{yin-etal-2020-tabert}, TUTA \cite{wang2021tuta}                                                     \\
WikiTableQuestion \cite{pasupat2015compositional} & \textbf{Task Related Content:} 22,033 question-answer pairs                                                                                                                                                                                   & TABLEFORMER \cite{yang2022tableformer}, TableQAKit \cite{lei2023tableqakit}, StructGPT \cite{jiang2023structgpt} \\
WikiSQL \cite{zhong2017seq2sql}                   & \textbf{Task Related Content}: 80,654 questions, answers and SQL queries                                                                                                                                                                      & TAPEX \cite{liu2021tapex}, TableQAKit \cite{lei2023tableqakit}, TAPAS \cite{herzig2020tapas}                     \\
Spider \cite{yu2018spider}                        & \textbf{Task Related Content:} 10,181 questions and 5,693 SQL queries                                                                                                                                                                         & TaBERT \cite{yin-etal-2020-tabert}, TUTA \cite{wang2021tuta}, StructGPT \cite{jiang2023structgpt}                \\
SQA \cite{iyyer2017search}                        & \textbf{Task Related Content:} 6,066 question sequences containing 17,553 questions-answer pairs                                                                                                                                              & TABLEFORMER \cite{yang2022tableformer}, GPT4Table \cite{sui2023gpt4table}, UniTabPT \cite{sarkar2023testing}     \\
MULTIHIERTT \cite{zhao2022multihiertt}            & {\textbf{Table Metadata:} hierarchical column headers and row headers\\\textbf{Context Information:} unstructured text\\\textbf{Task Related Content:} 10,440 question-answer pairs, annotations of reasoning processes and supporting facts} & MULTIHIERTT \cite{zhao2022multihiertt}, NAPG \cite{zhang2023napg}, TableQAKit \cite{lei2023tableqakit}           \\
FeTaQA \cite{nan2022fetaqa}                       & {\textbf{Table Metadata:} page title, section title\\\textbf{Task Related Content:} 10,330 question, free-form answer, and supporting table cells}                                                                                            & DATER \cite{ye2023large}, UniTabPT \cite{sarkar2023testing}                                                      \\
TAT-QA \cite{zhu2021tat}                          & {\textbf{Context Information:} 2,757 associated paragraphs\\\textbf{Task Related Content: }16,552 question-answer pairs with scale, derivation to arrive at the answer, and answer source}                                                    & TAT-QA \cite{zhu2021tat}, TableQAKit \cite{lei2023tableqakit}                                                    \\
HybridQA \cite{chen2020hybridqa}                  & {\textbf{Context Information: }293,269 hyperlinked passages\\\textbf{Task Related Content:} 69,611 question-answer pairs}                                                                                                                     & GPT4Table \cite{sui2023gpt4table}, TableQAKit \cite{lei2023tableqakit}                                           \\
ToTTo \cite{parikh2020totto}                      & {\textbf{Table Metadata:} page title, section title, section text\\\textbf{Task Related Content:} highlighted cells, final text (annotation)}                                                                                                 & GPT4Table \cite{sui2023gpt4table}, UniTabPT \cite{sarkar2023testing}                                             \\
TabFact \cite{chen2019tabfact}                    & \textbf{Task Related Content: }117,854 annotated statements                                                                                                                                                                                   & TAPEX \cite{liu2021tapex}, TABLEFORMER \cite{yang2022tableformer}, DATER \cite{ye2023large}                      
\end{talltblr}
\end{sidewaystable}

Furthermore, a special type of 2D tabular dataset incorporates dynamic temporal information, with some entries recorded over time along with their corresponding timestamps. Tables within the dataset covering various domains are linked by unique identifiers. For instance, MIMIC is a large publicly available electronic health record dataset, with its latest version being MIMIC-IV \cite{johnson2020mimic}.  It originates from the de-identified records of the Beth Israel Deaconess Medical Center, covering admissions from 2008 to 2019. MIMIC-IV adopts a modular structure, featuring three tabular schemas adopted relational structure: Hospital, ICU, and Emergency Department. These database schemas contain comprehensive patient information including demographics, laboratory measurements, medications, vital signs and more. Additionally, MIMIC-IV includes modules covering other medical data modalities: Note, Diagnostic Electrocardiogram, and Chest X-ray. UK Biobank is a large-scale prospective biomedical database comprising 500,000 individuals of middle and old age, recruited between 2016 and 2010 across the United Kingdom. It is a multimodal dataset containing an extensive amount of clinical information, such as patient demographics, detailed questionnaires, and various physical measurements. Additionally, the dataset also collects multimodal imaging data and genome-wide genotyping. The dataset is linked to electronic health records allowing for longitudinal follow-up to track health outcomes.

\section{Language Modeling Techniques on Tabular Data}
In this section, we analyze language model techniques and their adaptation to tabular data. Key components of tabular data modeling include input processing techniques, attention techniques, and training objectives.

\tikzset{%
    parent/.style =          {align=center,text width=2cm,rounded corners=3pt, line width=0.3mm, fill=gray!10,draw=gray!80},
    child/.style =           {align=center,text width=2.3cm,rounded corners=3pt, fill=blue!10,draw=blue!80,line width=0.3mm},
    grandchild/.style =      {align=center,text width=2cm,rounded corners=3pt},
    greatgrandchild/.style = {align=center,text width=1.5cm,rounded corners=3pt},
    greatgrandchild2/.style = {align=center,text width=1.5cm,rounded corners=3pt},    
    referenceblock/.style =  {align=center,text width=1.5cm,rounded corners=2pt},
    input_processing/.style =           {align=center,text width=1.8cm,rounded corners=3pt, fill=violet!10,draw=violet!80,line width=0.3mm},   
    input_processing_work/.style =           {align=center, text width=6cm,rounded corners=3pt, fill=violet!10,draw=violet!0,line width=0.3mm},   
    intermediate/.style =           {align=center,text width=1.8cm,rounded corners=3pt, fill= purple!10,draw= purple!80,line width=0.3mm},   
    intermediate_work/.style =           {align=center,text width=6cm,rounded corners=3pt, fill= purple!10,draw= purple!0,line width=0.3mm},     
    training_objectives/.style =           {align=center,text width=1.8cm,rounded corners=3pt, fill= teal!10,draw= teal!80,line width=0.3mm},   
    training_objectives_work/.style =           {align=center,text width=6cm,rounded corners=3pt, fill= teal!10,draw= teal!0,line width=0.3mm}, 
}

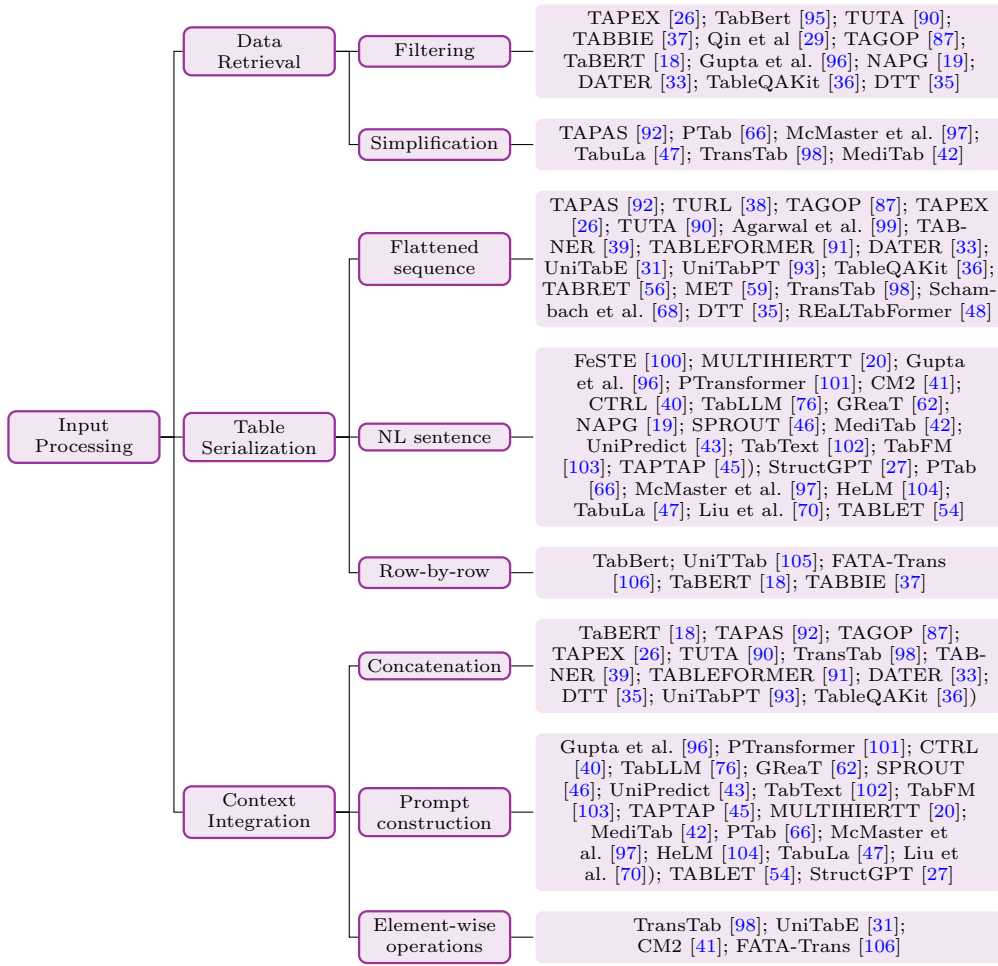
\begin{figure*}
\footnotesize
\centering
        \begin{forest}
            for tree={
                forked edges,
                grow'=0,
                draw,
                rounded corners,
                node options={align=center,},
                text width=2.7cm,
                s sep=6pt,
                calign=child edge, calign child=(n_children()+1)/2,
            },
            [Input Processing, input_processing
                [Data Retrieval, input_processing
                        [Filtering, input_processing
                            [TAPEX \cite{liu2021tapex}; TabBert \cite{padhi2021tabular}; TUTA \cite{wang2021tuta}; TABBIE \cite{iida2021tabbie}; Qin et al \cite{qin2022external}; TAGOP \cite{zhu2021tat}; TaBERT \cite{yin-etal-2020-tabert}; Gupta \text{et al.} \cite{gupta2022right}; NAPG \cite{zhang2023napg}; DATER \cite{ye2023large}; TableQAKit \cite{lei2023tableqakit}; DTT \cite{nobari2023dtt}, input_processing_work]
                        ]                    
                        [Simplification, input_processing
                            [TAPAS \cite{herzig2020tapas}; PTab \cite{liu2022ptab}; McMaster \text{et al.} \cite{mcmaster2023adapting}; TabuLa \cite{zhao2023tabula}; TransTab \cite{wang2022transtab}; MediTab \cite{wang2023meditab}, input_processing_work]  
                        ]
                ]
                [Table Serialization, input_processing
                        [Flattened sequence, input_processing
                            [TAPAS \cite{herzig2020tapas}; TURL \cite{deng2022turl};  TAGOP \cite{zhu2021tat}; TAPEX \cite{liu2021tapex};  TUTA \cite{wang2021tuta}; Agarwal et al\text{.} \cite{agarwal2022self}; TABNER \cite{koleva2022named}; TABLEFORMER \cite{yang2022tableformer}; DATER \cite{ye2023large}; UniTabE \cite{yang2023unitabe}; UniTabPT \cite{sarkar2023testing}; TableQAKit \cite{lei2023tableqakit}; TABRET \cite{onishi2023tabret}; MET \cite{majmundar2022met};  TransTab \cite{wang2022transtab}; Schambach et al\text{.} \cite{schambach2023scaling}; DTT \cite{nobari2023dtt}; REaLTabFormer \cite{solatorio2023realtabformer}, input_processing_work]
                            ]
                        [NL sentence, input_processing
                            [FeSTE \cite{harari2022few}; MULTIHIERTT \cite{zhao2022multihiertt}; Gupta et al\text{.} \cite{gupta2022right}; PTransformer \cite{ruan2024ptransformer}; CM2 \cite{ye2023ct}; CTRL \cite{li2023ctrl}; TabLLM \cite{hegselmann2023tabllm}; GReaT \cite{borisov2022language}; NAPG \cite{zhang2023napg}; SPROUT \cite{nam2023semi}; MediTab \cite{wang2023meditab}; UniPredict \cite{wang2023unipredict}; TabText \cite{carballo2023tabtext}; TabFM \cite{zhang2023towards}; TAPTAP \cite{zhang2023generative}); StructGPT \cite{jiang2023structgpt}; PTab \cite{liu2022ptab}; McMaster et al\text{.} \cite{mcmaster2023adapting}; HeLM \cite{belyaeva2023multimodal}; TabuLa \cite{zhao2023tabula}; Liu et al\text{.} \cite{liu2023investigating}; TABLET \cite{slack2023tablet}, input_processing_work]
                            ]
                        [Row-by-row, input_processing
                            [TabBert; UniTTab \cite{luetto2023one}; FATA-Trans \cite{zhang2023fata}; TaBERT \cite{yin-etal-2020-tabert}; TABBIE \cite{iida2021tabbie}, input_processing_work]
                        ]
                    ]
                [Context Integration, input_processing
                    [Concatenation, input_processing
                        [TaBERT \cite{yin-etal-2020-tabert}; TAPAS \cite{herzig2020tapas}; TAGOP \cite{zhu2021tat}; TAPEX \cite{liu2021tapex}; TUTA \cite{wang2021tuta}; TransTab \cite{wang2022transtab}; TABNER \cite{koleva2022named}; TABLEFORMER \cite{yang2022tableformer}; DATER \cite{ye2023large}; DTT \cite{nobari2023dtt}; UniTabPT \cite{sarkar2023testing}; TableQAKit \cite{lei2023tableqakit}), input_processing_work]]
                    [Prompt construction, input_processing
                        [Gupta et al. \cite{gupta2022right}; PTransformer \cite{ruan2024ptransformer}; CTRL \cite{li2023ctrl}; TabLLM \cite{hegselmann2023tabllm}; GReaT \cite{borisov2022language}; SPROUT \cite{nam2023semi}; UniPredict \cite{wang2023unipredict}; TabText \cite{carballo2023tabtext}; TabFM \cite{zhang2023towards}; TAPTAP \cite{zhang2023generative}; MULTIHIERTT \cite{zhao2022multihiertt}; MediTab \cite{wang2023meditab}; PTab \cite{liu2022ptab}; McMaster et al. \cite{mcmaster2023adapting}; HeLM \cite{belyaeva2023multimodal}; TabuLa \cite{zhao2023tabula}; Liu et al. \cite{liu2023investigating}); TABLET \cite{slack2023tablet}; StructGPT \cite{jiang2023structgpt}, input_processing_work]]
                    [Element-wise operations, input_processing
                        [TransTab \cite{wang2022transtab}; UniTabE \cite{yang2023unitabe}; CM2 \cite{ye2023ct}; FATA-Trans \cite{zhang2023fata}, input_processing_work]]
    ]
            ]
        \end{forest}
            \caption{The taxonomy of input processing. It contains data retrieval, table serialization and content integration.}
            \label{fig:input processing}
\end{figure*}

\subsection{Input Processing}
For transformer-based language models, it is typical to use text sequences as inputs. Prior to the standard embedding process, tabular data must be preprocessed through several steps to ensure compatibility with LM-based tabular models. These steps include data retrieval, table serialization, and context integration. Data retrieval plays a key role in certain frameworks, helping to maintain compliance with the language model's input size limits and boosting training efficiency. Table serialization involves modifying the original tabular data to fit the input specifications of language models. Context integration is also critical for effectively modeling tabular data, particularly when using pre-trained language models. The taxonomic overview of input processing is illustrated in Figure \ref{fig:input processing}.

\subsubsection{Data Retrieval}
The primary objective of the data retrieval module in LM-based tabular models is to expedite the training process, thereby cutting down on computational demands and training duration, especially when dealing with large tabular datasets. This module also plays a crucial role in ensuring that tabular inputs adhere to the token limits of language models. Typically, data retrieval strategies fall into two main categories.

\textbf{Filtering.} Filtering aims to reduce the number of rows or columns to be processed in large tables. Several straightforward techniques have been introduced for this purpose. For example, TAPEX \cite{liu2021tapex} employs random selection of rows to manage input size. TabBert \cite{padhi2021tabular} suggest using a sliding window to sample consecutive rows. TUTA \cite{wang2021tuta} split large tables into non-overlapping rows with same header, while TABBIE \cite{iida2021tabbie} and Qin et al \cite{qin2022external} truncate tables to a preset row and column limit to prevent memory issues.  TAGOP \cite{zhu2021tat} retains only those tables with number of rows and columns below a fixed threshold. 
Other methods, such as TaBERT \cite{yin-etal-2020-tabert}, Gupta et al. \cite{gupta2022right}, NAPG \cite{zhang2023napg}, DATER \cite{ye2023large}, TableQAKit \cite{lei2023tableqakit}, DTT \cite{nobari2023dtt}, incorporate specialized modules for extracting the most relevant top-n rows from tables.

\textbf{Simplification.} Simplification entails condensing the data by streamlining table contents. Some approaches simplify the content and context information such as TAPAS \cite{herzig2020tapas}, PTab \cite{liu2022ptab}, McMaster et al. \cite{mcmaster2023adapting}, TabuLa \cite{zhao2023tabula}. This might involve abbreviating or replacing column names and categorical values with synonymous terms. Additional methods, such as TransTab \cite{wang2022transtab}, MediTab \cite{wang2023meditab}, opt to omit negative binary features. This strategy is especially effective in reducing computational and memory requirements when dealing with inputs that have high-dimensional, sparse one-hot features.

\subsubsection{Table Serialization}
\label{serialization}
Table serialization involves converting tabular data into a format suitable for LM-based tabular systems. This process can be categorized into three different types.

\begin{figure}[!h]
    \centering
    \includegraphics[scale=0.58]{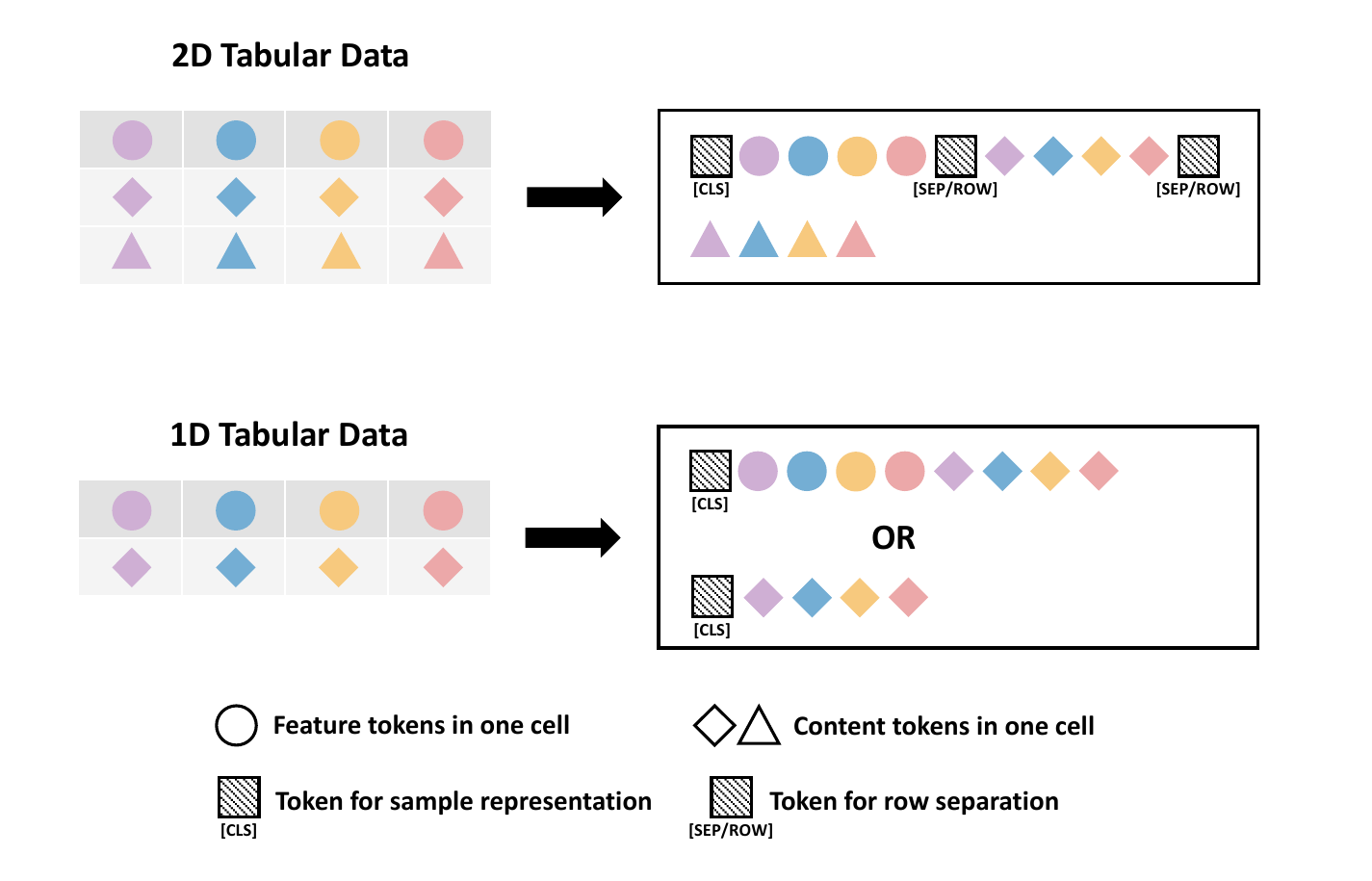}
    \caption{The illustration of flattened sequence for 1D (down) and 2D (up) data in table serialization.}
    \label{fig:flatten}
\end{figure}
\textbf{Flattened sequence.} The table is converted into flattened sequence, which transforms tabular data, either 1D or 2D, into a predefined flattened sequence format. This method is widely adopted due to its simplicity and directness. Figure \ref{fig:flatten} provides an overview of this method for table serialization.
More precisely, for 2D tables, the prevalent methodology involves reformatting these tables into a linear sequence of words pieces. This process integrates specific separators, such as [CLS] for representation generation, and [SEP] and [ROW] for row separation, to efficiently structure the data. Significant research in this domain includes TAPAS \cite{herzig2020tapas}, TURL \cite{deng2022turl},  TAGOP \cite{zhu2021tat}, TAPEX \cite{liu2021tapex},  TUTA \cite{wang2021tuta}, Agarwal et al. \cite{agarwal2022self}, TABNER \cite{koleva2022named}, TABLEFORMER \cite{yang2022tableformer}, DATER \cite{ye2023large}, UniTabE \cite{yang2023unitabe}, UniTabPT \cite{sarkar2023testing}, and TableQAKit \cite{lei2023tableqakit}. 
For tasks that require additional information besides tabular data, such as QA tasks, the tokens from the additional information (e.g., query or question) are usually positioned ahead of the flattened table-derived tokens.
In contrast, models designed for 1D tables often rely solely on representation generator to facilitate a streamlined sequence, as seen in TABRET \cite{onishi2023tabret}, MET \cite{majmundar2022met},  TransTab \cite{wang2022transtab}, Schambach et al. \cite{schambach2023scaling}, DTT \cite{nobari2023dtt}, and REaLTabFormer \cite{solatorio2023realtabformer}.

\textbf{Natural language (NL) sentence.} It is typically used with pre-trained language models to make full use of the textual information. This adaptation is essential for training and using pre-trained models, particularly in tasks such as masked language modeling, where understanding language senmantics is crucial. Many systems have designed the prompts for each feature to be transformed into natural sentences, and then concatenate them into a whole NL sentences for finetuning (FeSTE \cite{harari2022few}, MULTIHIERTT \cite{zhao2022multihiertt}, Gupta et al. \cite{gupta2022right}, PTransformer \cite{ruan2024ptransformer}, CM2 \cite{ye2023ct}, CTRL \cite{li2023ctrl}, TabLLM \cite{hegselmann2023tabllm}, GReaT \cite{borisov2022language}, NAPG \cite{zhang2023napg}, SPROUT \cite{nam2023semi}, MediTab \cite{wang2023meditab}), UniPredict \cite{wang2023unipredict}, TabText \cite{carballo2023tabtext}, TabFM \cite{zhang2023towards}, TAPTAP \cite{zhang2023generative}), and StructGPT \cite{jiang2023structgpt}. 
Alternative methodologies have been explored to achieve this objective by simplifying language prompts with separators (PTab \cite{liu2022ptab}, McMaster et al. \cite{mcmaster2023adapting}, HeLM \cite{belyaeva2023multimodal}, TabuLa \cite{zhao2023tabula}, Liu et al. \cite{liu2023investigating}, TABLET \cite{slack2023tablet}). These transformed formats, despite their modifications, continue to align with the domain of NL sentences.

\textbf{Row-by-row.} It involves processing data on a row-by-row basis, typically employed for handling data where each row represents a essential entity, with an emphasis on exploring the interactions among various rows. This approach is predominantly utilized in longitudinal tabular data, where each row symbolizes the attributes at a specific timestamp. Processing longitudinal tabular data in this manner aids in uncovering both intra and inter-temporal information. For example, TabBert \cite{padhi2021tabular} introduced a field transformer to discern local associations within a single record at a given time step, while its sequence transformer captures broader relationships across different time steps, thus addressing the temporal aspects of the data. UniTTab \cite{luetto2023one} builds upon TabBert's framework by incorporating a linear projection layer, enabling the processing of rows with varying internal structures and types within time series. Based on TabBert, FATA-Trans \cite{zhang2023fata} designs a unique approach to separately process static and dynamic fields, integrating time interval information through a time-aware position embedding that operates on a row-by-row basis.

Moreover, some models have adopted row-by-row table serialization to align with their architectural designs, extending beyond just longitudinal tabular data application. For instance, TaBERT \cite{yin-etal-2020-tabert} introduces vertical self-attention, which operates across vertically aligned encoded vectors from different rows. Similarly, TABBIE \cite{iida2021tabbie} employs a row Transformer to encode cells along each row of the table, complemented by a column Transformer that performs an analogous function across columns, thereby facilitating the generation of contextualized cell embeddings.

\subsubsection{Context Integration}
In LM-based tabular data modeling, the integration of table context with its content is a pivotal step. 

\textbf{Concatenation.} The most straightforward technique involves the sequential combination of the context with table data, typically achieved through serial concatenation. A notable example is TaBERT \cite{yin-etal-2020-tabert}, which merges column names, types, and cell values in table serialization with the [SEP] symbol as a separator. Alternatively, some approaches prefer concatenating all column names prior to all cell values as illustrated in Figure (TAPAS \cite{herzig2020tapas},  TAGOP \cite{zhu2021tat}, TAPEX \cite{liu2021tapex}, TUTA \cite{wang2021tuta}, TransTab \cite{wang2022transtab}, TABNER \cite{koleva2022named}, TABLEFORMER \cite{yang2022tableformer}, DATER \cite{ye2023large}, DTT \cite{nobari2023dtt}, UniTabPT \cite{sarkar2023testing}, TableQAKit \cite{lei2023tableqakit}).

\textbf{Prompt construction.} Another commonly adopted technique of context integration utilizes prompt construction, often in conjunction with pre-trained language models on tabular datasets. This approach often entails the transformation of context data into prompts, augmented with additional words or separators. Some methods implements the combination of column names with connective words to NL sentences (Gupta et al. \cite{gupta2022right}, PTransformer \cite{ruan2024ptransformer}, CTRL \cite{li2023ctrl}, TabLLM \cite{hegselmann2023tabllm}, GReaT \cite{borisov2022language}, SPROUT \cite{nam2023semi}, UniPredict \cite{wang2023unipredict}, TabText \cite{carballo2023tabtext}, TabFM \cite{zhang2023towards}, TAPTAP \cite{zhang2023generative}). Additionally, MULTIHIERTT \cite{zhao2022multihiertt} employs a Facts Retrieving Module to convert table metadata into NL context sentences, while MediTab \cite{wang2023meditab} leverages LLMs to transform tabular data into NL sentences, utilizing column names as the contextual information. For simplicity, some models such as PTab \cite{liu2022ptab}, McMaster et al. \cite{mcmaster2023adapting}, HeLM \cite{belyaeva2023multimodal}, TabuLa \cite{zhao2023tabula}, Liu et al. \cite{liu2023investigating}),TABLET \cite{slack2023tablet}, StructGPT \cite{jiang2023structgpt}, directly concatenate column names with cell values, using colons or spaces as contextual separators during the pre-training phase of the model.

\textbf{Element-wise operations}. Furthermore, there are approaches that incorporate context through element-wise operations such as addition and multiplication. For instance, TransTab \cite{wang2022transtab} implements element-wise product between column name embeddings and cell value embeddings for binary and numerical features. UniTabE \cite{yang2023unitabe} introduces a unique linking layer, designed to intricately fuse information from column names with their corresponding cell values. Similarly, CM2 \cite{ye2023ct} utilizes an element-wise multiplication approach, combining normalized numerical values with their respective header embeddings. FATA-Trans \cite{zhang2023fata} adopts an element-wise summation technique, combining record embeddings, field type embeddings, and time-aware position embeddings to generate sequence embeddings with the sequential encoding transformer. 

Apart from these methods, some models have developed specialized modules for context integration. For example, TABBIE \cite{iida2021tabbie} concatenates column names with cell values using row and column transformers. Qin et al. \cite{qin2022external} enhance large-scale tabular pre-training models by integrating a common-sense knowledge graph, providing a flexible and easily integrable solution.

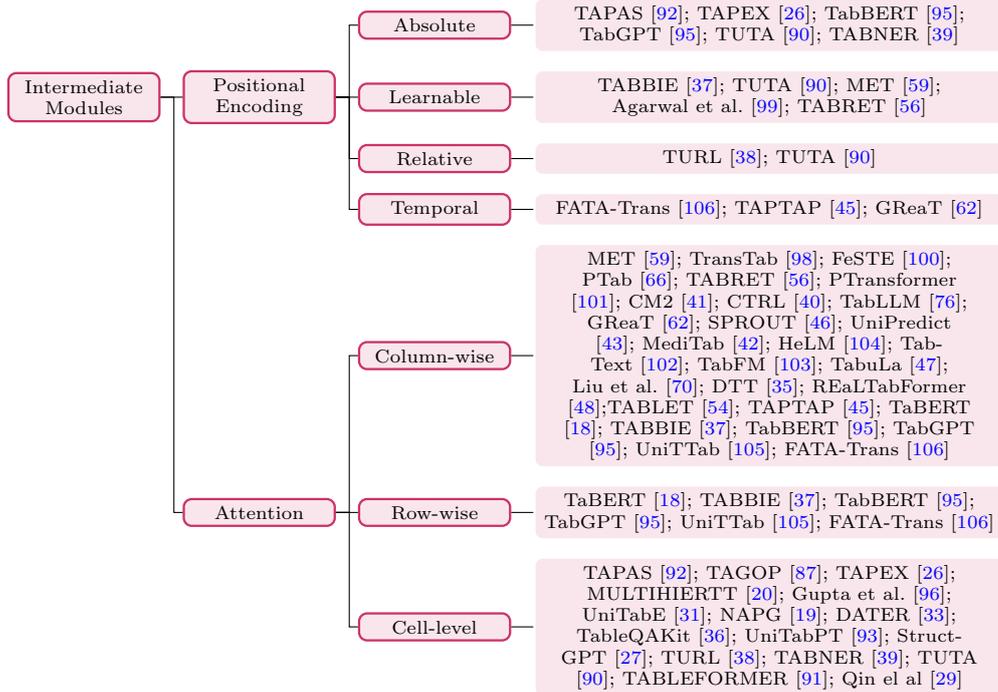
\begin{figure*}
\footnotesize
\centering

        \begin{forest}
            for tree={
                forked edges,
                grow'=0,
                draw,
                rounded corners,
                node options={align=center,},
                text width=2.7cm,
                s sep=6pt,
                calign=child edge, calign child=(n_children()+1)/2,
            },
            [Intermediate Modules, intermediate
            [Positional Encoding, intermediate
                [Absolute, intermediate
                    [TAPAS \cite{herzig2020tapas}; TAPEX \cite{liu2021tapex}; TabBERT \cite{padhi2021tabular}; TabGPT \cite{padhi2021tabular}; TUTA \cite{wang2021tuta}; TABNER \cite{koleva2022named}, intermediate_work]
                ]
                [Learnable, intermediate     
                    [TABBIE \cite{iida2021tabbie}; TUTA \cite{wang2021tuta}; MET \cite{majmundar2022met}; Agarwal et al. \cite{agarwal2022self}; TABRET \cite{onishi2023tabret}, intermediate_work ]
                ]
                [Relative, intermediate
                    [TURL \cite{deng2022turl}; TUTA \cite{wang2021tuta}, intermediate_work]
                ]
                [Temporal, intermediate
                    [FATA-Trans \cite{zhang2023fata}; TAPTAP \cite{zhang2023generative}; GReaT \cite{borisov2022language}, intermediate_work]
                ]
            ]
            [Attention, intermediate
                [Column-wise, intermediate
                    [MET \cite{majmundar2022met}; TransTab \cite{wang2022transtab}; FeSTE \cite{harari2022few}; PTab \cite{liu2022ptab}; TABRET \cite{onishi2023tabret}; PTransformer \cite{ruan2024ptransformer}; CM2 \cite{ye2023ct}; CTRL \cite{li2023ctrl}; TabLLM \cite{hegselmann2023tabllm}; GReaT \cite{borisov2022language}; SPROUT \cite{nam2023semi}; UniPredict \cite{wang2023unipredict}; MediTab \cite{wang2023meditab}; HeLM \cite{belyaeva2023multimodal}; TabText \cite{carballo2023tabtext}; TabFM \cite{zhang2023towards}; TabuLa \cite{zhao2023tabula}; Liu et al. \cite{liu2023investigating}; DTT \cite{nobari2023dtt}; REaLTabFormer \cite{solatorio2023realtabformer};TABLET \cite{slack2023tablet}; TAPTAP \cite{zhang2023generative}; TaBERT \cite{yin-etal-2020-tabert}; TABBIE \cite{iida2021tabbie}; TabBERT \cite{padhi2021tabular}; TabGPT \cite{padhi2021tabular}; UniTTab \cite{luetto2023one}; FATA-Trans \cite{zhang2023fata}, intermediate_work]
                ]
                [Row-wise, intermediate     
                    [TaBERT \cite{yin-etal-2020-tabert}; TABBIE \cite{iida2021tabbie}; TabBERT \cite{padhi2021tabular}; TabGPT \cite{padhi2021tabular}; UniTTab \cite{luetto2023one}; FATA-Trans \cite{zhang2023fata}, intermediate_work ]
                ]
                [Cell-level, intermediate
                    [TAPAS \cite{herzig2020tapas}; TAGOP \cite{zhu2021tat}; TAPEX \cite{liu2021tapex}; MULTIHIERTT \cite{zhao2022multihiertt}; Gupta et al. \cite{gupta2022right}; UniTabE \cite{yang2023unitabe}; NAPG \cite{zhang2023napg}; DATER \cite{ye2023large}; TableQAKit \cite{lei2023tableqakit}; UniTabPT \cite{sarkar2023testing}; StructGPT \cite{jiang2023structgpt}; TURL \cite{deng2022turl}; TABNER \cite{koleva2022named}; TUTA \cite{wang2021tuta}; TABLEFORMER \cite{yang2022tableformer}; Qin el al \cite{qin2022external} , intermediate_work]
                ]
            ]]
        \end{forest}
            \caption{The taxonomy of intermediate modules. It contains positional encoding and attention mechanism.}
            \label{fig:intermediate_modules}
\end{figure*}

\subsection{Intermediate Modules}
In addition to input processing techniques, researchers are also interested in modifying intermediate modules within transformer architectures to better adapt them to the tabular domain. As illustrated in Figure \ref{fig:intermediate_modules}, this includes two major components: positional encoding and attention mechanisms.

\subsubsection{Positional Encoding}
Positional encoding serves as a fundamental technique within deep learning methodologies, particularly evident in the architecture of transformers, facilitating the model's ability to capture crucial ordering information in input sequences. This necessity arises from the distinct processing mechanism of transformers compared to recurrent neural networks (RNNs) or long short-term memory networks (LSTMs), which inherently process data sequentially. It's worthy noting that transformers require a method to integrate positional information into the frameworks. Therefore, some adaptations have been proposed to incorporate additional positional encoding for rows or columns to explicitly capture the underlying table structure.

\textbf{Absolute.} The first form of positional encoding, known as absolute positional encoding, was initially employed in the original transformer model architecture to inject information about token positions within a sequence. It comprises sinusoidal functions of different frequencies to encode each position, enabling the model to understand the sequence's order. This approach has also been adapted for tabular data modeling to retain information concerning the tabular structure, as observed in works such as TAPAS \cite{herzig2020tapas}, TAPEX \cite{liu2021tapex}, TabBERT, and TabGPT \cite{padhi2021tabular}. TUTA \cite{wang2021tuta} utilized tree-based positional encodings to better capture hierarchical information. However, in TABNER \cite{koleva2022named}, the authors noted that, positional encoding across the entire table could blur the Named Entity Recognition (NER) training signal for BERT.

\textbf{Learnable.} Unlike the fixed mathematical formula used in absolute encoding, learned positional encodings comprise learnable parameters that the model can update during training. This enables the model to learn an embedding for each position, potentially capturing more intricate patterns than the absolute encoding. For instance, TABBIE \cite{iida2021tabbie} and TUTA \cite{wang2021tuta} introduce various tree-based positional embedding approaches tailored to the hierarchical nature of tabular data, with parameters initialized randomly and tuned through training. MET \cite{majmundar2022met}, Agarwal et al. \cite{agarwal2022self}, and TABRET \cite{onishi2023tabret} develop learnable positional encodings to encode column-specific information.

\textbf{Relative.} Relative positional encoding differs from absolute encoding by encoding the relative positions of tokens in input sequences rather than their absolute positions. It allows the model to focus on the distance between tokens in each modality, where the relative positioning of elements holds greater significance than their absolute position. Techniques such as those employed in TURL \cite{deng2022turl}, which provides a positional embedding vector containing relative position information for a token within captions or headers, and TUTA \cite{wang2021tuta}, which incorporates in-cell position encoding to encode single tokens relative to their cell positions, exemplify the application of relative positional encoding within various contexts.

\textbf{Temporal.} In practical settings involving the modeling of longitudinal tabular data, temporal information such as timestamps is available and informative. Despite their informational richness, these temporal information is often underutilized in conventional approaches. In addressing this limitation, FATA-Trans \cite{zhang2023fata} introduces a novel approach: the exploration of learnable time-aware position embedding. This innovative technique considers both the sequence order and time intervals between rows, enabling the model to effectively understand underlying temporal patterns within a sequence.

Notably, tabular data exhibits permutation invariance, implying that its structure remains unchanged despite arbitrary rearrangements of columns. Consequently, certain frameworks in the tabular domain opt to forgo positional encoding to mitigate the introduction of positional bias, particularly when modeling 1D tabular data \cite{ye2023ct, huang2020tabtransformer}. For some models that utilize natural language sentences as a table serialization method, permutation functions are proposed to shuffle the order of features. This approach aims to mitigate the impact of implicit injection on spurious positional relationships within textual encoding \cite{zhang2023generative, borisov2022language}.

\subsubsection{Attention}
Attention mechanisms in language modeling have significantly advanced the field of NLP by allowing models to dynamically focus on different parts of the input data when producing an output. This capability has been particularly impactful in dealing with sequential data, such as text, where understanding the context and the relationships between words or tokens is crucial for tasks like translation, summarization, and question-answering.

Tabular data, characterized by its structured format in rows and columns, presents a challenge in applying attention mechanisms, compared to sequential text data. However, the concept of attention can still be beneficial in this context, especially for tasks that involve understanding relationships between different rows or columns within the table.

\textbf{Column-wise.}
Column-wise attention mechanisms are dominantly used in the analysis of tabular datasets as it helps to understand the data structure by capturing the relationship between columns. This could be useful in tasks where the output depends on understanding complex interactions between features. 
In the domain of 1D tabular data analysis, researchers have proposed to directly use stacked transformer to incorporate attention mechanisms on tabular data, as illustrated in the works MET \cite{majmundar2022met}, TransTab \cite{wang2022transtab}, FeSTE \cite{harari2022few}, PTab \cite{liu2022ptab}, TABRET \cite{onishi2023tabret}, PTransformer \cite{ruan2024ptransformer}, CM2 \cite{ye2023ct}, CTRL \cite{li2023ctrl}. These models leverage the strengths of transformer architectures to capture complex feature interactions effectively. Further innovations have emerged through the integration of LLMs, focusing on serialized tabular data and NL prompts. Key advancements include TabLLM \cite{hegselmann2023tabllm}, GReaT \cite{borisov2022language}, SPROUT \cite{nam2023semi}, UniPredict \cite{wang2023unipredict}, MediTab \cite{wang2023meditab},HeLM \cite{belyaeva2023multimodal}, TabText \cite{carballo2023tabtext}, TabFM \cite{zhang2023towards}, TabuLa \cite{zhao2023tabula}, Liu et al. \cite{liu2023investigating}, DTT \cite{nobari2023dtt}, REaLTabFormer \cite{solatorio2023realtabformer}, TABLET \cite{slack2023tablet}, TAPTAP \cite{zhang2023generative}. These studies highlight the potential of leveraging attention mechanisms on tabular data, opening new avenues for column-wise feature interaction and processing of tabular information.

For 2D tabular data, particularly with complex tasks such as question answering and information retrieval, the deployment of attention mechanisms also plays a pivotal role in facilitating the aggregation of information across column-wise features and, when applicable, across different modalities. For example, TaBERT \cite{yin-etal-2020-tabert} leverages attention in transformer to to bridge the gap in relationship exploration between questions and relevant rows. Similarly, TABBIE \cite{iida2021tabbie} utilizes row transformers to compute row embeddings, thereby enabling the generation of column-wise contextualized cell embeddings.
Moreover, in handling longitudinal tabular data, the advent of field transformers marks a significant advancement by being attentive on column-wise features at individual timestamps to extract valuable information, as demonstrated in TabBERT, TabGPT \cite{padhi2021tabular} and UniTTab \cite{luetto2023one}. FATA-Trans \cite{zhang2023fata} introduces both static and dynamic field transformers to cater to varying column-wise data types, enhancing its utility across a broad spectrum of tasks using longitudinal tabular data.

\textbf{Row-wise.}
To explore relationships between multiple rows within tabular data, particularly for tasks like time series prediction, the multi-head attention mechanism emerges as a crucial tool by weighting the importance of different rows. A notable implementation of this is the vertical attention mechanism proposed in TaBERT \cite{yin-etal-2020-tabert}, which aggregates information across diverse rows in a content snapshot, enabling TaBERT to effectively capture cross-row dependencies on cell values. Furthermore, TABBIE \cite{iida2021tabbie} introduces a column transformer with a row transformer to produce row-wise contextualized representations that enrich the model's interpretability.
For longitudinal tabular data, sequence encoding transformer serves as a fundamental framework on the top of field transformer to effectively integrate time-sensitive information across multiple rows through the attention mechanisms ( $e.g.$, TabBERT, TabGPT \cite{padhi2021tabular}, UniTTab \cite{luetto2023one}, and FATA-Trans \cite{zhang2023fata}.)

\textbf{Cell-level.}
Attention can also be applied at the cell level, where the model learns to focus on specific cells within the table that are most relevant for the task at hand. This approach can be particularly useful for dealing with heterogeneous data or data with high-dimensional features for complex downstream tasks such as question answering, information retrieval.
Many methodologies leverage BERT \cite{devlin2018bert} or its variants \cite{liu2019roberta, lewis2019bart, reimers2019sentence} to implement attention on table cells within 2D tabular data through pre-training (TAPAS \cite{herzig2020tapas},  TAGOP \cite{zhu2021tat}, TAPEX \cite{liu2021tapex}, MULTIHIERTT \cite{zhao2022multihiertt}, Gupta et al. \cite{gupta2022right}, UniTabE \cite{yang2023unitabe}, NAPG \cite{zhang2023napg}) or via PLM/LLM (DATER \cite{ye2023large}, TableQAKit \cite{lei2023tableqakit}, UniTabPT \cite{sarkar2023testing}, StructGPT \cite{jiang2023structgpt}). Notably, beyond employing multi-head self-attention, TURL \cite{deng2022turl} and TABNER \cite{koleva2022named} introduce a visibility matrix as an attention mask, enabling the information aggregation only on structurally related tokens during self-attention calculation. Further advancements include TUTA \cite{wang2021tuta}, which incorporates structure-aware tree-based attention to capture spatial and hierarchical information in tables, and TABLEFORMER \cite{yang2022tableformer}, which enhances table understanding and alignment with text by introducing task-independent relative attention biases.

In addition to the previous attention mechanisms on tabular data, some work has developed cross-attention mechanism that integrate information from table and other modalities. For instance, Qin el al \cite{qin2022external} have proposed path-wise attention layer to align the cross-domain representation with the weighted contribution on tabular data and external knowledge graph.

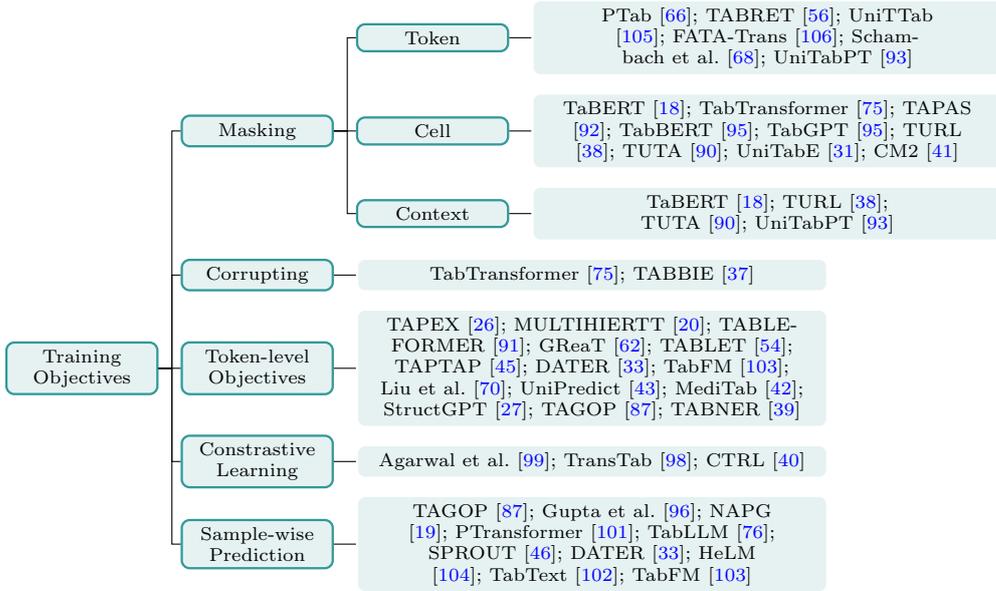
\begin{figure*}
\footnotesize
\centering
        \begin{forest}
            for tree={
                forked edges,
                grow'=0,
                draw,
                rounded corners,
                node options={align=center,},
                text width=2.7cm,
                s sep=6pt,
                calign=child edge, calign child=(n_children()+1)/2,
            },
            [Training Objectives, training_objectives
                [Masking, training_objectives
                    [Token, training_objectives
                        [PTab \cite{liu2022ptab}; TABRET \cite{onishi2023tabret}; UniTTab \cite{luetto2023one};  FATA-Trans \cite{zhang2023fata}; Schambach et al. \cite{schambach2023scaling}; UniTabPT \cite{sarkar2023testing}, training_objectives_work]
                    ]
                    [Cell, training_objectives
                        [TaBERT \cite{yin-etal-2020-tabert}; TabTransformer \cite{huang2020tabtransformer}; TAPAS \cite{herzig2020tapas}; TabBERT \cite{padhi2021tabular}; TabGPT \cite{padhi2021tabular}; TURL \cite{deng2022turl}; TUTA \cite{wang2021tuta}; UniTabE \cite{yang2023unitabe}; CM2 \cite{ye2023ct}, training_objectives_work]
                    ]
                    [Context, training_objectives
                        [TaBERT \cite{yin-etal-2020-tabert}; TURL \cite{deng2022turl}; TUTA \cite{wang2021tuta}; UniTabPT \cite{sarkar2023testing}, training_objectives_work]
                    ]
                ]
                [Corrupting, training_objectives    
                    [TabTransformer \cite{huang2020tabtransformer}; TABBIE \cite{iida2021tabbie}, training_objectives_work ]
                ]
                [Token-level Objectives, training_objectives
                    [TAPEX \cite{liu2021tapex}; MULTIHIERTT \cite{zhao2022multihiertt}; TABLEFORMER \cite{yang2022tableformer}; GReaT \cite{borisov2022language}; TABLET \cite{slack2023tablet}; TAPTAP \cite{zhang2023generative}; DATER \cite{ye2023large}; TabFM \cite{zhang2023towards}; Liu et al. \cite{liu2023investigating}; UniPredict \cite{wang2023unipredict}; MediTab \cite{wang2023meditab}; StructGPT \cite{jiang2023structgpt};  TAGOP \cite{zhu2021tat}; TABNER \cite{koleva2022named}, training_objectives_work]
                ]
                [Constrastive Learning, training_objectives
                    [Agarwal et al. \cite{agarwal2022self}; TransTab \cite{wang2022transtab}; CTRL \cite{li2023ctrl}, training_objectives_work]
                ]
                [Sample-wise Prediction, training_objectives
                    [TAGOP \cite{zhu2021tat}; Gupta et al. \cite{gupta2022right}; NAPG \cite{zhang2023napg}; PTransformer \cite{ruan2024ptransformer}; TabLLM \cite{hegselmann2023tabllm}; SPROUT \cite{nam2023semi}; DATER \cite{ye2023large}; HeLM \cite{belyaeva2023multimodal}; TabText \cite{carballo2023tabtext}; TabFM \cite{zhang2023towards}, training_objectives_work]
                ]
            ]
        \end{forest}
            \caption{The taxonomy of training objectives. It includes masking, corrupting, token-level objectives, contrastive learning and sample-wise prediction.}
            \label{fig:training_objectives}
\end{figure*}

\subsection{Training Objectives}
Training objectives in language modeling are designed to help the model learn the semantics of natural language, enabling it to perform well on various downstream tasks. In tabular data modeling, the training objectives are typically modified to be adapted into the structure of tabular data. An overview of these training objectives for language modeling on tabular data is illustrated in Figure \ref{fig:training_objectives}.
\subsubsection{Masking}
The Masked Language Model (MLM) is firstly introduced in the pre-training phase of BERT \cite{devlin2018bert}. In the MLM, some tokens within a sequence of text are randomly masked, and the model is trained to predict these masked tokens based on the surrounding context. This task helps the model understand the contextual relationships between tokens.

\textbf{Token.} Several studies \cite{liu2022ptab, onishi2023tabret, luetto2023one, zhang2023fata, schambach2023scaling, sarkar2023testing} have adopted the token-level masking as their main training objective, which involves randomly masking a specific percentage of tokens with [MASK] to facilitate the acquisition of contextual representations from text-based tabular data.

\textbf{Cell.} In contrast to conventional NLP pre-training by masking a random subset of tokens, several studies \cite{yin-etal-2020-tabert, huang2020tabtransformer, herzig2020tapas, padhi2021tabular, deng2022turl, wang2021tuta} have embraced the approach of masking whole cells and training models to predict these masked cells. Building upon this strategy of whole cell masking, CM2 \cite{ye2023ct} introduces the utilization of column names as prompts to enhance whole cell masking effectiveness in cross-table pre-training.

\textbf{Context.} Beyond the focus on understanding table contents through token-level and cell-level masking strategies, table context—including metadata, surrounding texts, and even query questions—plays a critical role in tabular data modeling for complex tasks alongside table contents. For example, TaBERT \cite{yin-etal-2020-tabert} incorporates an MLM objective with a 15\% masking rate for sub-tokens in NL contexts and a Masked Column Prediction objective to facilitate model learning of column names and types. TURL \cite{deng2022turl} and TUTA \cite{wang2021tuta} utilize an MLM objective to enable models to grasp the lexical, semantic, and contextual information within table metadata for cell understanding. UniTabPT \cite{sarkar2023testing} also employs the MLM objective on NL/SQL texts and table headers. It's noted that instead of masking several tokens in header, the model masks all tokens in single header to bolster the model's capability in recovering column information from context, thereby deepening its understanding of the relationship between table column values and their headers.

\subsubsection{Corrupting}
Apart from the conventional MLM-based training objective, corrupting emerges as another potent strategy for tabular representation learning. This approach is particularly beneficial for tabular data, where detecting corrupted cells is crucial for accurate table structure decomposition tasks, in which incorrectly identified row/column separators or cell boundaries can result in corrupted cell contents. For example, TabTransformer \cite{huang2020tabtransformer} implements this by replacing features with random values and utilizing a binary classifier to determine feature corruption. TABBIE \cite{iida2021tabbie} further introduces two corruption methods: frequency-based cell sampling and intra-table cell swapping, with the former showing significant effectiveness across most downstream table-based tasks.


\subsubsection{Token-level Objectives}

Token-level objectives include two main categories, the foremost being seq2seq (sequence-to-sequence) learning. This approach involves training models to produce text sequences based on given inputs, proving especially beneficial for machine translation, summarization, and question answering applications. Existing studies \cite{liu2021tapex, zhao2022multihiertt, yang2022tableformer, borisov2022language, slack2023tablet, zhang2023generative, ye2023large, zhang2023towards, liu2023investigating, wang2023unipredict, wang2023meditab, jiang2023structgpt} have explored the application of seq2seq learning objectives within the tabular domain. This exploration involves reinterpreting traditional tasks, such as classification, as question answering task on tables. In addition to predicting the tokens conditioned on previous input, models can further be tailored to predict a variety of token-level characteristics, including but not limited to part-of-speech tags, named entities, syntactic configurations, and classifications at the token level. An illustrative case is TABNER \cite{koleva2022named}, which validates the utility of language models for addressing a unique named entity recognition challenge within industrial spreadsheets. Similarly,  TAGOP \cite{zhu2021tat} adeptly extracts supporting evidence from a hybrid context, containing both tables and related texts, through sequence tagging. This approach also facilitates the introduction of a Number Order Classifier \cite{zhu2021tat}, aimed at determining the sequential positioning of two numerals in the final outcome.

\subsubsection{Contrastive Learning}
Contrastive learning in language models brings representations of similar text sequences closer and pushes dissimilar ones apart. In tabular data modeling, this approach is adapted for various purposes. Agarwal et al. \cite{agarwal2022self} have introduced an additional contrastive loss component to mask prediction loss, treating embeddings from different masked instances of the identical user as positive examples. Furthermore, TransTab \cite{wang2022transtab} presents a novel vertical-partition contrastive learning approach, which substantially increases the scope of positive and negative samples for learning more informative sample representation. CTRL \cite{li2023ctrl} takes tabular and corresponding textual data as two modalities, employing contrastive learning for knowledge alignment and integration.

\subsubsection{Sample-wise Prediction}
Sample-wise prediction involves the training of models through the utilization of sample representations, commonly the [CLS] token, for the predictions of various tasks. A plethora of studies \cite{ruan2024ptransformer,hegselmann2023tabllm,nam2023semi,ye2023large,belyaeva2023multimodal,carballo2023tabtext,zhang2023towards, gupta2022right} that leveraged PLMs/LLMs, which do not require extensive training corpora, have adopted this approach for training their models across diverse downstream tasks. Moreover,  TAGOP \cite{zhu2021tat} introduces an Operator Classifier that accounts for the correct scale in addition to the accurate number. Concurrently, NAPG \cite{zhang2023napg} has unveiled a non-autoregressive program generation framework equipped with five predictors based on sample representations for numerical reasoning. This innovation enables the parallel generation of programs, significantly enhancing efficiency in numerical reasoning tasks.

\section{Evolution of How Language Models are Adapted}
In addition to providing foundational concepts and modeling techniques, this survey emphasizes the evolution of language model adaptation for tabular data and explores potential future research directions. As illustrated in Figure \ref{fig:timeline}, tabular modeling initially benefited from language modeling in two major ways: pre-training from scratch, where researchers focused on critical elements (e.g., embeddings, attention) within language models to improve decision-making, and using PLMs, where some researchers integrated PLMs as modules within larger models to leverage prior semantic knowledge for addressing more complex tasks. In recent years, the field of NLP has advanced into the era of LLMs, marking a series of paradigm shifts that have introduced novel capabilities and enhanced the efficiency of tabular data processing.

\subsection{Pre-training From Scratch and PLMs}
Prior to the emergence of LLMs, tabular modeling predominantly followed two distinct approaches based on the objectives.
\subsubsection{Pre-training From Scratch}
The first approach includes models trained from scratch, primary designed for tabular prediction tasks. To improve their ability to extract relevant features and correlations from tables, researchers predominantly adopt transformer architectures, integrating fundamental language modeling techniques. This included modifications to embedding methods and training objectives to better handle tabular data.
For instance, Table2Vec \cite{zhang2019table2vec} learns meaningful embeddings for table elements such as captions, column headings, and cells, aiding tasks such as row population and table retrieval. 
TabTransformer \cite{huang2020tabtransformer} uses Transformer architectures with a self-attention mechanism to convert categorical feature embeddings into strong contextual embeddings, which are combined with continuous features to boost prediction accuracy. This model is pre-trained with multiple objectives, including masked language modeling and replaced token detection, making it robust against noisy and missing data. 
CM2 \cite{ye2023ct} introduces an efficient cross-table pre-training framework featuring a semantic-aware tabular model that uniformly encodes heterogeneous tables with minimal restrictions. It utilizes a novel pre-training objective called prompt masked table modeling, enabling scalable pre-training on diverse tables. 
Schambach et al. \cite{schambach2023scaling} propose a Transformer-based architecture for cross-table representation learning, employing table-specific tokenizers and a shared Transformer backbone to minimize inductive biases. This approach includes both single-table and cross-table models, trained using a self-supervised masked cell recovery objective, improving the model's ability to learn representations across different tables.

Moreover, researchers are adapting transformer-based architecture to different applications including table understanding, time series analysis and tabular prediction.
%
Table understanding represents an advanced task in tabular data modeling, with several models leveraging the transformer architecture to enhance comprehension of table structures. 
For example, TAGOP \cite{zhu2021tat} introduces an innovative question-answering model based on RoBERTa that reasons over both tables and text. It uses sequence tagging to extract relevant cells from tables and text spans, applying symbolic reasoning with aggregation operators to derive answers.
On the other hand, TUTA \cite{wang2021tuta} provides a unified pre-training architecture for structured tables, utilizing tree-based attention and positional embeddings to effectively capture both spatial and hierarchical information. The model incorporates three pre-training objectives: masked language modeling, cell-level cloze tasks, and table context retrieval, which collectively facilitate token, cell, and table-level representations. Pre-trained on a vast array of diverse web and spreadsheet tables, TUTA is subsequently fine-tuned for tasks such as cell and table type classification.
TURL \cite{deng2022turl} introduces a structure-aware Transformer encoder that models relational tables using a Masked Entity Recovery objective, effectively capturing semantics and knowledge from large-scale unlabeled data, and demonstrating strong performance across six table understanding tasks. 
Similarly, TABLEFORMER \cite{yang2022tableformer} presents a robust structure-aware table-text encoding architecture that incorporates learnable attention biases to mitigate biass from table linearization, outperforming strong baselines in numerical experiments on datasets like SQA, WTQ, and TABFACT.

In addition to aforementioned tasks on 2D tabular data, time series data is special type of 2D tabular data, on which recent advancements in transformer-based architectures have revolutionized the modeling. 
TabBERT and TabGPT \cite{padhi2021tabular} are pioneers in this field, with TabBERT enabling end-to-end pre-training for classification or regression tasks, and TabGPT generating realistic synthetic tabular sequences while preserving patient privacy. 
FATA-Trans \cite{zhang2023fata} innovatively utilizes two field transformers to distinguish static and dynamic fields and employs time-aware position embeddings to capture temporal patterns. 
Agarwal et al. \cite{agarwal2022self} introduce a self-supervised transformer model that learns from both sequential and tabular features, excelling in tasks such as supervised classification and click bot detection without labels. 
UniTTab \cite{luetto2023one}, on the other hand, focuses on heterogeneous time-dependent data, using continuous embedding vectors for numerical and categorical features, and is uniformly trained with a masked token task, showing robust performance across various tasks. 
Additionally, UniTabE \cite{yang2023unitabe} presents a versatile framework for handling tables, addressing pre-training challenges on large-scale tabular data using free-form prompts to encode inputs into a Transformer encoder, with a curated dataset of approximately 13 billion samples from Kaggle aiding its pre-training phase.
 
Tabular prediction utilizing 1D tabular data has significantly benefited from language modeling techniques, resulting in enhanced prediction performance. For instance, MET \cite{majmundar2022met} proposes a reconstruction-based approach for tabular representation learning using masked encoding to reconstruct original input data through a series of stacked transformers.
TransTab \cite{wang2022transtab} introduces a transferable tabular Transformer architecture that converts each data sample into a generalizable embedding vector, processed through stacked Transformers for feature encoding, incorporating column descriptions and table cells into a gated transformer model. It uses both supervised and self-supervised pre-training methods to boost performance.
TABRET \cite{onishi2023tabret}, a pre-trainable Transformer-based model, adapts to unseen columns in downstream tasks with an extra retokenizing step before fine-tuning, calibrating feature embeddings based on masked autoencoding loss.

While these methods have demonstrated effectiveness in specific contexts, they often required considerable amounts of tabular data for pre-training from scratch in order to perform well for specific tasks. This issue is especially pronounced in data-sensitive fields such as healthcare, where the availability of extensive, high-quality data is often limited, and privacy concerns are critical.

\subsubsection{Pre-trained Language Models}
The second approach involves the use of PLMs within tabular modeling. These models are especially suited for tasks requiring a deeper semantic understanding, such as TQA. PLMs like BERT are favored in this approach due to their foundational pre-trained architectures, which require less training data and yield superior predictive performance compared to earlier methods. Additionally, these pre-trained models allow for fine-tuning on task-specific datasets, enhancing both the efficiency and effectiveness of the modeling process.

For the TQA task,
TaBERT \cite{yin-etal-2020-tabert} proposes pre-trained language model built on top of Bert to simultaneously process textual and tabular data. It uses content snapshots and vertical self-attention mechanisms to capture the association between table content and natural language. 
TAPAS \cite{herzig2020tapas} is a weakly supervised question answering model that extends BERT’s masked language model objective to structured data. It answers questions by selecting relevant cells and performing aggregation operations, bypassing the need for generating logical forms. Employing a unique pre-training strategy, TAPAS was further trained on millions of tables and associated text segments sourced from Wikipedia. 
%
TAPEX \cite{liu2021tapex} develops an innovative table pre-training method that simulates a neural SQL executor to learn structured tabular data effectively. It overcomes the scarcity of high-quality tabular data by utilizing synthetic SQL queries and their execution results as a pre-training corpus. 
%
MULTIHIERTT \cite{zhao2022multihiertt} is a benchmark specifically crafted for complex numerical reasoning tasks involving documents with multiple hierarchical tables and textual content, primarily derived from financial reports. The dataset is distinguished by several features: it includes documents with multiple tables and extensive unstructured texts, predominantly hierarchical tables, and necessitates intricate reasoning more challenging than current benchmarks. It also provides detailed annotations of reasoning paths and supporting facts. To address the challenges in MULTIHIERTT, a novel question-answering model, MT2NET is proposed. It uses the RoBERTa model as encoder and is the first one that conducts fact retrieval to pinpoint relevant facts before engaging a reasoning module for symbolic processing. 
%
TableQuery \cite{abraham2022tablequery} introduces a tool for querying tabular data with natural language, overcoming limitations of existing deep learning methods for question answering on tabular data. TableQuery utilizes pre-trained LM for free-text question answering to convert natural language queries into structured queries, avoiding the need to load large datasets into memory or serialize databases. It supports various column types and does not require re-training, allowing for easy integration of better-performing models.
%
NAPG \cite{zhang2023napg} proposes a non-autoregressive program generation framework based on the RoBERTa encoder, specifically designed for numerical reasoning in mixed tabular-textual question answering scenarios. It uniquely generates complete program tuples, including both operators and operands, independently, which effectively mitigates the exposure bias problem seen in traditional autoregressive models and drastically increases generation speed.

PLMs are also used for other tasks such as TP, for example, 
FeSTE \cite{harari2022few} is a Transformer-based framework based on Bert and designed to enrich tabular datasets by leveraging unstructured data. It trains on various datasets to create versatile models suitable for additional datasets using few-shot learning. FeSTE introduces a fine-tuning method that converts dataset tuples into sentences and uses pre-trained LM to generate valuable features from external data sources effectively. 
PTab \cite{liu2022ptab} proposes a framework that utilizes pre-trained LM (Bert) to enhance the contextual representation of tabular data, enabling training on mixed datasets. The framework processes tabular data in three stages: Modality Transformation (MT), Masked Language Fine-Tuning (MF), and Classification Fine-Tuning (CF). 
P-Transformer \cite{ruan2024ptransformer} devises a prompt-based multimodal transformer architecture tailored for medical tabular data. This framework comprises two key components: a tabular cell embedding generator based on RoBERTa and a tabular transformer. The embedding generator effectively encodes inputs from both structured and unstructured tabular data into a unified language semantic space, utilizing a pre-trained sentence encoder and specialized medical prompts. The transformer component then integrates these cell representations to create comprehensive patient embeddings, which are applied across diverse medical tasks.
%
CTRL \cite{li2023ctrl} is a framework that enhances Click-Through Rate (CTR) prediction by integrating collaborative and language models. It transforms tabular data into text format, which is then processed using both a collaborative model and a pre-trained LM (RoBERTa). This approach allows for the alignment of collaborative and semantic signals via cross-modal learning. 

In addition to TQA and TP, researchers have also demonstrated that PLMs enhance performance in other semantically demanding tasks. For example,
TABNER \cite{koleva2022named} proposes a table transformer model tailored for the industrial Named Entity Recognition (NER) challenge, designed to identify entities in complex, structured spreadsheets. It incorporates a domain-specific data augmentation technique that uses knowledge graphs to enhance performance in low-resource environments and address the technical complexities of industrial data. The research underscores the importance of tabular inductive bias for model convergence and shows that this data augmentation method markedly boosts performance compared to sequential models.
%
TABBIE \cite{iida2021tabbie} introduces a self-supervised model tailored for tabular data, employing a corrupt cell detection objective to learn table structure and semantics. It initializes cell embeddings using BERT and utilizes two distinct Transformers to separately encode rows and columns. TABBIE leverages the self-supervised ELECTRA objective for pre-training and excels in various downstream tasks such as column population, row population, and column type prediction.
Qin et al. \cite{qin2022external} proposes dual-adapters within a pre-trained tabular Transformer model to bridge domain discrepancies between external knowledge sources and tabular data. These dual-adapters operate in parallel: one adapter is trained on knowledge graph triplets, while the other processes semantically enhanced tables. Furthermore, a path-wise attention layer is integrated to enhance the cross-domain representation. 
%
Gupta et al. \cite{gupta2022right} explores enhancing NLP systems' interpretability and reliability through Trustworthy Tabular Inference. This task ensures systems substantiate their predictions with clear evidence. Employing a two-stage approach, the methodology begins by extracting evidence from tables, then uses this data to predict inference labels.

\subsection{Emergence of Large Language Models}\label{llms}

Large language models have recently achieved remarkable success in the domain of natural language understanding.
These LLMs often have billions of parameters and pre-trained on enormous text corpus, leading to strong zero-shot or few-shot capability on a variety of tasks such as reasoning, question answering, and code generation.
Hence, it naturally raises the question of how can we leverage the advantage of LLMs to push the boundaries of tabular data modeling.

In contrast to most \textbf{Pre-training From Scratch} and \textbf{PLMs} which are designed for specific tasks, researchers have begun to explore the use of a single unified model for a broader and more challenging range of tabular tasks, such as few-shot classification and table generation.
%

In the Table Prediction domain,
TabLLM \cite{hegselmann2023tabllm} explore LLMs' proficiency in zero-shot and few-shot classification tasks specifically applied to tabular data. The performance of LLMs is notably sensitive to the specific details of the natural language input. Therefore, TabLLM studied nine different serialization techniques and found that simple text template ($e.g.$, the [\textit{column\_name}] is [\textit{value}]) perform best in almost all datasets for zero-shot and few-shot classifications. Following serialization, TabLLM uses a short task description as a prompt to obtain output probabilities from the LLM and then finetune the LLM. 
%
TAPTAP \cite{taptap} proposes table pre-training for tabular prediction, where the model was pre-trained on 450 tabular datasets with a total of nearly 2 million samples. TAPTAP pre-train the model by prediction the subsequent token using serialized tables. With pre-training, TAPTAP is capable of generating high-quality synthetic tables, which can support various applications involving tabular data. 
SPROUT \cite{nam2023semisupervised} extends the application of LLMs to semi-supervised tabular data modeling. It first provides a LLM with a selection of labeled samples and prompts it to identify the most important feature for the downstream task. In order to maintain the relevance of the constructed examples and the real downstream task, SPROUT generates prompts that predict the selected important features based upon the remaining column features. The generated prompts are then combined with descriptions of a few labeled samples to further bolster the in-context-learning performance of the LLM. 
%
UniPredict \cite{wang2023unipredict} presents an universal predictor for tabular data classification tasks. It aimed to create a more adaptable tabular model, wherein a single set of parameters could be applied universally across datasets from any domain. To achieve this goal, UniPredict integrates the prompts for serialized tables and target confidence from external predictors to finetune its backbone LLM. 
MediTab \cite{wang2023meditab} focuses on medical tabular data where existing models are often trained on a single dataset for a specific task, leading to poor generalizibility. It uses a LLM to consolidate the tabular data: by describing one sample in a variety of ways, MediTab can generates diverse consolidated samples as data augmentation. Meanwhile, MediTab employed an audit module to prevent potential hallucinations from LLM. To benefit from the out-domain datasets, MediTab expands the available data for one task by aligning a trained model on all other different tasks to obtain supplementary data. 
HeLM \cite{belyaeva2023multimodal} presents a multimodal LLM for health domain with the use of individual specific data. Unlike other tabular models in health domain, HeLM integragtes non-tabular data such as medical time sereis into the serialization process. Specifically, it trains separate encoders for non-texutal data modality on top of LLM to learn their mapping to the same representation space as text. HeLM has shown effectiveness for zero/few-shot disease prediction on UK biobank dataset. However, it was also observed that the tuned model degraded in conversational ability.
%
TabText \cite{carballo2023tabtext} leverages the text embeddings of the serialized table as additional contexutal information to improve the performance of standard machine learning performance. An significant advantage of TabText is that the LLM is flexible to replace when new models available. 
TABLET\cite{slack2023tablet} is a benchmark designed to assess the ability of LLMs to learn from instructions for tabular prediction tasks. It encompasses 20 different tabular datasets, each annotated with instructions that differ in phrasing, granularity, and technicality. TABLET enables researchers to gauge model performance in tabular predictions based on in-context instructions alone (the zero-shot setting) or with a limited number of labeled examples (the few-shot setting). Interestingly, the findings indicate that while instructions generally enhance LLM performance, LLMs exhibit a strong bias against accurately classifying certain instances and do not consistently follow the provided in-context instructions.
%
Liu et al. \cite{liu2023investigating} investigates the fairness issues associated with employing pre-trained LLMs for tabular prediction tasks. They found that LLMs, specifically GPT-3.5 turbo, often depend on social biases present in their pre-training data for making predictions in tabular tasks. While few-shot in-context learning can somewhat reduce these biases, it does not completely remove them. In addition, the gap in fairness metrics across different subgroups remains larger compared to traditional machine learning models, such as Random Forest and shallow Neural Networks. 
%
TabFM \cite{zhang2023towards} introduces a method for training a LLM based foundation model specifically for tabular data. It involves incorporating examples from 115 tabular datasets, which are serialized into text for the training process. TabFM aims to create a model that effectively generalizes to new tabular datasets in both zero and few-shot settings. The results reveal that their training approach lead to better performance on tabular tasks compared to other models like GPT4 and TabLLM. Moreover, when trained with an increased number of data points, TabFM achieves performance comparable to traditional tabular models such as XGBoost.

Meanwhile, by leveraging the powerful language generation capabilities of LLMs, researchers have used these models to generate more complex and realistic tables.
GReaT \cite{borisov2022language} introduces a novel approach for the modeling and generation of realistic, heterogeneous tabular data, leveraging the capabilities of LLMs. Contrary to conventional tabular data generation techniques which convert the data into a purely numerical format, GReaT encodes tabular data into textual representations, effectively capturing the underlying semantics. GReaT implements a random feature order permutation on the serialized tabular data and finetunes the pretrained LLM on samples devoid of order dependencies. This approach enables arbitrary conditioning in the generation of tabular data. In the generation phase, GReaT feeds the LLMs a text description, prompting them to proceed with completion. 
REaLTabFormer \cite{solatorio2023realtabformer} proposes a generative model for relational tabular data generation. REaLTabFormer contains a partent table model and a child table model to generate synthetic observations and synthetic related observations, respectively. It uses a GPT-2 model to generate a parent table, and subsequently produce the child tables that conditioned on the parent table using a GPT-2 decoder. 
%
TabuLa \cite{zhao2023tabula} proposes several approaches to enhance tabular data synthesis. Unlike traditional methods that rely on pre-trained models, TabuLa utilizes a randomly initialized model, enabling quicker adaptation to the specific requirements of tabular data synthesis tasks. Furthermore, TabuLa compresses all column names and categorical values into a single token each, effectively reducing sequence length. It also implements a middle padding strategy, ensuring that features within the same data column retain their absolute positions in the encoded token sequence, thus preserving the integrity of the original data structure.
%
DTT \cite{nobari2023dtt} explores the challenge of converting tabular data from one source format to another target format using only a few examples. The approach begins by decomposing the problem into smaller subtasks and serializing the input. It then employs a LLM (ByT5) to predict outcomes for each subtask. An aggregator combines these individual predictions to produce a final output. Experimental evaluations indicate that the performance of DTT matches or surpasses that of other LLMs such as GPT3, despite notable differences in size. 
Furthermore, DTT also enhances the performance of LLMs in the table transformation task by integrating them into it.

Additionally, researchers have focused on harnessing the strong reasoning capabilities of LLMs for table understanding tasks such as TQA. For example,
Dater \cite{dater} explores the application of LLMs for table-based reasoning tasks. It focuses on addressing two main issues associated with table-based reasoning: (i) LLMs are unable to directly encoding large tables due to input token limits; (ii) direct decompose complex questions could easily fall into a hallucination dilemma. Dater tackles the two main challenges by leveraging the in-context learning in LLMs. It first decompose a large table into a small table that relevant to the question  using LLM alongside a handful of prompting examples. Following this, Dater decompose a complex question into simpler step-by-step sub-questions. Notably, Dater operates without task-specific finetuning that might diminish the LLMs' in-context capabilities. 
%
GPT4Table \cite{sui2023gpt4table} establishes a benchmark aimed at studying which factors of input design most significantly influence LLMs' ability to understand tabular data. The proposed Structural Understanding Capabilities (SUC) benchmark includes seven tasks, each designed to compare different input configurations. The findings indicate that while LLMs possess fundamental abilities in interpreting tabular data, correctly choosing the input design could be as a key element in enhancing the SUC. Therefore, authors proposed a self-augmented prompting method to provide additional knowledge and constraints, thus enhancing the LLM’s performance in downstream tasks. 
Chain-Of-Table \cite{wang2024chainoftable} has further augmented the reasoning abilities of LLMs by utilizing the tabular structure to generate intermediate thought processes during table-based reasoning tasks. It guides LLMs to dynamically create a sequence of operations in response to a given table and its related question, offering more accurate and reliable predictions. 

The emergence of these LLM-based tabular modeling methods represents a paradigm shift from models pre-trained from scratch or PLMs designed for specific tasks to LLM-based methods capable of handling various tasks. This shift opens up new opportunities and challenges for future exploration.

\section{Challenges and Future Opportunities}
As discussed in Section \ref{llms},
LLMs has already been used in many tabular data applications, such as predictions, data synthesis, question answering and table understanding. 
Here we outline some practical limitations and considerations for future
research.
Specially, we discuss four challenges for language modeling on tabular data, consisting of computation efficiency, interpretability, biases, and data types.
Computation efficiency and interpretability are general challenges in AI application scenarios with PLMs and LLMs. 
Biases is also a general concern, but this problem has different meaning when focus on tabular data.
Data types is a particular challenge for tabular data. For other LLMs/PLMs application scenarios, it is rare to simultaneously face many different data type, such as numerical, categorical, binary, text, timestamps, and even nested data.

\subsection{Computation Efficiency}

Compared to methods based on data mining or PLMs, using LLMs significantly require more computational resources. 
This increase in computational resources is not only a cost issue but also a practical challenge in scenarios where sufficient hardware support is not available, such as when mobile devices are required.
To address this issue, there are two directions worth exploring.
The first solution is that how to improve the efficiency of training and inference for large models. 
There are numerous studies focused on enhancing the efficiency of LLMs, such as the use of Adapter modules~\cite{he2021towards, houlsby2019parameter, hu2023llm}. 
This technique involves integrating smaller neural network modules into the intermediate layers of PLMs or LLMs.
Another method is Prefix Tuning~\cite{li2021prefix, he2023virtual}, where a trainable prefix is added either to the input sequence or to the hidden layers to facilitate more efficient training.
LoRA~\cite{hu2021lora} represents an advanced strategy for achieving parameter-efficient fine-tuning while circumventing common issues associated with other methods. 
The fundamental principle of LoRA is to approximate the parameter updates of a full-rank weight matrix using a low-rank matrix.
However, these methods, while speeding up the training process, do not provide help when perform inference.
Meanwhile, deploying such a large model remains challenging.

The second direction is that leveraging more powerful LLMs to enhance the performance of smaller models to approach larger ones. 
For instance, the studies~\cite{ho2022large, hsieh2023distilling, li2022explanations} let a LLM act as a teacher and distil knowledge into a small model.
Besides, some studies~\cite{abdullin2024synthetic, schmidhuber2024llm} employ LLMs to generate synthetic data to further enhance smaller models.
However, knowledge distillation faces several challenges, including the inability to precisely control the information being distilled, a tendency for overfitting, and high sensitivity to the quality of data. When using synthetic data, while it can enhance the scale of model training, it also carries the risk of introducing biases which may skew the results. Furthermore, evaluating synthetic data presents additional complexities. 
For such reason, addressing these issues requires further exploration of more effective methods to enhance computational efficiency.

\subsection{Interpretability}

Over the last decade, interest in interpretability has surged, driven by the proliferation of large datasets and the advancement of deep neural networks. Despite LLMs demonstrating exceptional capabilities across diverse applications, the depth of interpretability research remains relatively superficial.
Research on neural network interpretability can be categorized into two distinct groups. The first focuses on result interpretability. For instance, when a neural model is asked a question such as, "There are 5 apples on the table. If Aria used 2 to make dinner and then added 10 more apples on the table, how many apples are on the table?" a typical model might simply respond, "There are 13 apples on the table".
This response lacks an explanation of the reasoning process. However, with the introduction of Chain of Thought (CoT) technology~\cite{wei2022chain}, models are now capable of explaining their reasoning in a step-by-step manner, such as "Starting with 5 apples, 2 were used, leaving 3. Adding the 10 apples gives a total of 13." This approach significantly enhances interpretability by allowing models to articulate the thought process behind their conclusions, essentially using the models themselves to explain their reasoning. Recent advancements in retrieval argument generation (RAG) and the integration of knowledge graph represent similar efforts to bolster interpretability.

The second research stream theoretically explores the internal mechanisms of LLMs to shed light on their operations. For example, the study by Ansuini et al.\cite{ansuini2019intrinsic} utilizes the concept of intrinsic dimensionality to analyze network layers. It reveals that the intrinsic dimensionality of each layer in a trained network is significantly smaller than the number of units in each layer, compared to an untrained network. This phenomenon aids in assessing the generalization performance of the network and enhances our understanding of neural network interpretability. 
Another study\cite{deng2023understanding} employs various attribution methods to determine the relevance or contribution of each input variable to the final output. 
Despite these efforts, most studies in this area face strong constraints or focus only on specific aspects, making a generalized approach to interpreting neural networks a continuing challenge.

\subsection{Biases}

LLMs often inherit biases from their training data, affecting their fairness in tasks like TQA, TR, and TMP. 
For instance, Liu et al.~\cite{liu2023investigating} assessed the fairness of GPT-3.5 in tabular predictions using few-shot learning. 
The findings confirmed that biases in training data significantly contribute to model biases.
Additionally, Mao et al.~\cite{mao2022biases} showed that PLMs and LLMs might produce biased outcomes based on task design, prompt crafting, and label word selection. The study highlighted how variations in label definitions and arrangements can affect model effectiveness and introduce biases.

For tabular data modeling, serialization introduces specific biases. 
Common methods like flattening the table sequence, converting tables into natural language sentences, or employing row-by-row serialization all result in some loss of table structure information, and make tables into plain input sequences.
However, according to Liu et al.~\cite{liu2024lost}, LLMs tend to focus more on the beginning and end of the input sequence, often overlooking the middle sections, which complicates the retrieval of relevant information. 
Each serialization approach also impacts how much attention each cell receives, introducing potential biases.

Given these issues, exploring more effective serialization methods is crucial. 
Moreover, addressing the inherent biases of LLMs deserves more focused attention to enhance model fairness and efficacy.

\subsection{Data Types}

Diverse data types is one of the most challenging issue in language modeling tabular data. 
In other fields and tasks, it is rare to simultaneously handle numerical, categorical, binary, text, hyperlinks, timestamps data, and even these data exist nested scenario.
Traditionally, data mining-based approaches perform well in handling multiple data types, by developing specific modeling methods for each type. However, these approaches significantly lag behind LLM-based methods in tasks involving text understanding, generation, and reasoning. Consequently, more research is shifting towards LLM-based methods for addressing tabular data issues. However, LLMs are not naturally suited for handling multiple data types, especially numerical data. While some studies~\cite{zhao2022multihiertt,ye2023ct,hegselmann2023tabllm} have circumvented this issue by converting numeric data into text, this solution significantly increases the model's input length, introducing new problems.

Currently, there are LLMs studies~\cite{taptap,belyaeva2023multimodal} specifically trained on table data, which somewhat enhance the model's ability to understand various data types. Nevertheless, the following issues remain worthy of further exploration:
1) During training, how to balance the proportion of different types of data in large-scale table data, maintaining data integrity while ensuring the diversity and scale of training data.
2) How to design more appropriate unsupervised pre-training tasks. Unlike traditional language models, table data may involve different levels and granularities of attention interactions between rows, columns, and the entire table, and sometime even existed nested tables. Under such condition, various data types present additional challenges.

\section{Conclusion}
Language models for tabular data capitalize on innovative techniques to harness the intricacies of heterogeneous data structures, thereby enabling analytics across diverse applications. 
This comprehensive survey delves into the core aspects of language modeling for tabular data from various dimensions, including foundational structures, methodologies, and the evolution of modeling techniques. 
A detailed taxonomy and analysis of techniques, ranging from data retrieval to the strategic use of intermediate modules and training objectives, offer a granular view of the field's current state. 
The evolution of language modeling is critically reviewed in two pivotal stages: (i) Initial explorations involving bespoke pre-training adjustments for tabular contexts and the integration of pre-trained models like BERT for enhanced semantic understanding. (ii) The recent shift towards leveraging LLMs represents a significant paradigm shift, broadening the scope for addressing more complex tabular data challenges. 
This paper not only highlights the substantial progress made but also articulates four major challenges and proposes potential avenues for future research in this promising and rapidly evolving area, making it a clear roadmap for navigating future explorations in language modeling for tabular data.

\section*{Declarations}

\subsubsection*{Funding}
This work is supported by the Cisco-NUS Accelerated Digital Economy Corporate Laboratory (Award I21001E0002, funded by A*STAR, CISCO Systems (USA) Pte. Ltd, and National University of Singapore).
\subsubsection*{Competing interests} 
The authors declare no conflict of interest.
\subsubsection*{Data availability} No datasets were generated or analysed during the current study.

\subsubsection*{Author contribution} 
Y.R. \& X.L.: Conceptualization, Methodology, Resources, Data Acquisition, Writing(Original Draft Preparation), Writing(Review \& Editing), Visualization; 
J.M. \& Y.D.: Resources, Data Acquisition, Writing(Original Draft Preparation); 
K.H.: Writing(Original Draft Preparation); 
M.F.: Writing(Review \& Editing), Supervision, Funding Acquisition





\bibliography{sn-bibliography}

\end{document}